\documentclass[runningheads]{llncs}
\usepackage[T1]{fontenc}
\usepackage{graphicx}
\usepackage{booktabs}
\usepackage{amsmath}
\usepackage{amsfonts}
\usepackage[misc]{ifsym}

\usepackage{tabularx}
\usepackage{multirow}
\usepackage{hhline}

\usepackage{graphicx}
\usepackage{amssymb}
\usepackage{amsmath}
\usepackage{mathtools}
\usepackage[table,xcdraw]{xcolor}
\usepackage{multirow}
\usepackage{lineno}
\usepackage{enumitem}
\usepackage{bbding}
\usepackage{float} 

\usepackage{mwe}

\usepackage{xcolor}
\usepackage{enumitem}
\usepackage{adjustbox}
\usepackage{amsmath}

\newcommand{\todo}[1]{{\color{red}[TODO: #1]}}

\newcommand{\h}[4]{h_{#1}^{#2}({#3},{#4})}

\newcommand{\trainClean}{\mathbf{X}_\mathtt{tr}^0}
\newcommand{\calibClean}{\mathbf{X}_\mathtt{ca}^0}
\newcommand{\testClean}{\mathbf{X}_\mathtt{te}^0}

\newcommand{\testAnom}{\mathbf{X}_\mathtt{te}^1}
\newcommand{\calibAnom}{\mathbf{X}_\mathtt{ca}^1}

\newcommand{\ncf}{\mathtt{ncf}}

\begin{document}

\title{Localized Anomaly Detection\\ via Differentiable D-vine Copulas}

\titlerunning{Localized Anomaly Detection via Differentiable D-vine Copulas}

\author{Nicholas A. Pearson\inst{1}\Envelope\orcidID{0009-0008-4285-6831} \and Francesca Zanello \inst{2} \and \newline Davide Russo\inst{2} \and Luca Bortolussi\inst{1}\orcidID{0000-0001-8874-4001} \and \newline Francesca Cairoli\inst{1}\orcidID{0000-0002-6994-6553}}

\authorrunning{}

\institute{University of Trieste, Trieste, Italy\newline \and Idrostudi srl, Trieste, Italy\newline\email{nicholasandrea.pearson@phd.units.it}}

\maketitle              

\begin{abstract}

Vine copulas provide a flexible framework for modeling complex multivariate distributions through a hierarchical decomposition into bivariate pair-copulas. Fitting a D-vine requires selecting a copula family and parameter configuration for each pair-copula from a set of candidates encoding different dependence patterns. As the number of variables and candidate families increases, the number of possible configurations grows combinatorially. Existing fitting procedures address this challenge through sequential greedy decisions, committing to a single locally optimal family at each step and potentially discarding configurations that would yield a better global fit. To overcome this limitation, we propose a novel estimation framework that combines gradient-based maximum likelihood estimation, enabled by our fully differentiable implementation, with a beam-search strategy that maintains multiple competing D-vine configurations throughout the fitting process. This allows a broader exploration of the configuration space while remaining computationally tractable. Building on the fitted D-vine, we introduce a localized anomaly detection framework that exploits the hierarchical decomposition to produce both global anomaly scores and edge-level explanations. Statistical guarantees are provided through Mondrian conformal prediction, while the pair-copula structure enables the localization of anomalies to specific variable relationships. We evaluate the proposed framework on both benchmark and real-world datasets, demonstrating its effectiveness for interpretable anomaly detection with uncertainty quantification.

\end{abstract}

\section{Introduction}\label{sec:intro}
Anomaly Detection can be viewed as the task of identifying observations that deviate from the pattern exhibited by ordinary data. It plays a prominent role in a wide range of real-world applications such as fraud detection, industrial monitoring and healthcare~\cite{chandola2009anomaly}. In these settings, the consequences of incorrect identification can be significant, making it essential not only to predict if an observation is anomalous, but also to quantify the reliability of the corresponding predictions. It is often valuable to localize each anomaly, identifying the variables responsible for it and providing actionable insight into its underlying cause. To address these needs, we approach the task at hand as a two-stage problem. First, we estimate the joint distribution of the observations under ordinary conditions to obtain a model of normal behavior through copula modeling. Second, we build on this distribution, treating anomaly detection as a binary classification task, and employ a classifier whose predictions offer guarantees on their reliability through Conformal Prediction~\cite{angelopoulos2023conformal}.

Copulas provide a flexible framework for modeling the joint distribution of a set of variables by separating marginal behavior from the dependence structure~\cite{sklar1959fonctions}. Standard multivariate copulas scale poorly to high-dimensions~\cite{nelsen2006introduction}, a limitation overcome by pair-copula constructions~\cite{aas2009pair}, which decompose the multivariate distributions into a hierarchy of bivariate copulas. Out of these, we focus on the D-vine, which arranges the variables along a path ordering~\cite{czado2019analyzing}. Fitting a D-vine requires selecting a family and parameters for each pair-copula, a combinatorial problem that existing methods address either by exploring too few configurations~\cite{dissmann2013selecting} or at high computational cost~\cite{chapon2023imputation}.

 We propose a methodology that combines an efficient D-vine fitting procedure with a conformal anomaly detection scheme providing per-class reliability guarantees. Our contributions are:
 \begin{enumerate}[label=\roman*.]
    \item A fully differentiable PyTorch implementation of bivariate copula families, supporting gradient-based parameter estimation and GPU acceleration.
    \item A beam search procedure for D-vine fitting that, at each pair-copula, evaluates all candidate families and propagates forward only the most promising configurations. Several candidate paths are evaluated simultaneously, allowing for a broader exploration of the configuration space without incurring the cost of exhaustive search.
    \item An anomaly detection procedure framing detection as a classification task between ordinary and anomalous behavior. Mondrian conformal prediction is applied to obtain prediction regions that are guaranteed to contain the true label with a prescribed confidence level for each class. The hierarchical structure of D-vines is then exploited to localize each anomaly to individual pair-copulas in the decomposition and thus to the responsible variables.
\end{enumerate}

\noindent \textbf{Related Works.} Several approaches have been proposed for \emph{fitting D-vines}, with the standard approach of~\cite{dissmann2013selecting} selecting a single configuration in a greedy procedure which limits the number of explored configurations. Bayesian alternatives like the RJMCMC procedure of~\cite{chapon2023imputation} fully explore the configuration space but suffer from high computational costs and convergence issues. Examples of \emph{copula based anomaly detection} include CoCAI~\cite{pearson2025cocai} which combines copula modeling with conformal prediction for time series, and COPOD~\cite{li2020copod} which uses empirical copulas for outlier detection. Lastly,~\cite{horvath2020copula} investigates anomalous dependence patterns through unconditional pair-copulas defined on a spanning-tree structure.

\section{Background}

\subsection{Copula Theory}

Let $X=(X_1,\dots,X_d)$ be a random vector with joint cumulative distribution function (CDF) $F$ and continuous marginal CDFs $F_1,\dots,F_d$. By the probability integral transform, the variables $U_i = F_i(X_i)$ with $i=1,\dots,d,$ are uniformly distributed on $[0,1]$~\cite{yan2007enjoy}. The joint distribution of $(U_1,\dots,U_d)$ is called a \emph{copula and characterizes the dependence structure of $X$ independently of its marginals}~\cite{nelsen2006introduction}.
The central result of copula theory is Sklar's theorem~\cite{sklar1959fonctions}.

\begin{theorem}[Sklar's Theorem]
Let $F$ be a $d$-dimensional CDF with continuous marginals $F_1,\dots,F_d$. Then there exists a unique copula $C:[0,1]^d \to [0,1]$ such that
\begin{equation}
F(x_1,\dots,x_d)
=
C\left(F_1(x_1),\dots,F_d(x_d)\right).
\label{eq:sklarExistence}
\end{equation}
Conversely, for any copula $C$ and marginal CDFs $F_1,\dots,F_d$, Equation~\eqref{eq:sklarExistence} defines a valid $d$-dimensional distribution.
\end{theorem}
Sklar's theorem separates marginal behavior from dependence. If $C$ is absolutely continuous, its copula density is defined as
\begin{equation}
c(u_1,\dots,u_d)
=
\frac{\partial^d C(u_1,\dots,u_d)}
{\partial u_1 \cdots \partial u_d}.
\label{eq:copulaDensity}
\end{equation}
If the marginals are absolutely continuous with densities $f_1,\dots,f_d$, the joint density factorizes as
\begin{equation}
f(x_1,\dots,x_d)
=
c\left(F_1(x_1),\dots,F_d(x_d)\right)
\prod_{i=1}^{d} f_i(x_i).
\label{eq:joint-density}
\end{equation}
This decomposition forms the basis of likelihood-based inference and, in particular, of pair-copula constructions~\cite{aas2009pair}.

\subsubsection{Pair-Copula Constructions.}
Direct modeling with high-dimensional copulas is often impractical due to limited flexibility and restrictive dependence assumptions~\cite{min2010bayesian}. Pair-copula constructions (PCCs) address this by decomposing a multivariate copula into a collection of bivariate copulas, enabling flexible and scalable modeling of complex dependence structures~\cite{aas2009pair}.
Bivariate copulas model the dependence between two variables, and are also referred to as pair-copulas. They are typically classified into the elliptical and Archimedean families. Elliptical copulas, such as the Gaussian and Student-t copulas, are characterized by symmetric dependence patterns. In contrast, Archimedean copulas, including the Gumbel and Clayton copulas, capture asymmetric and tail-dependent behaviors~\cite{dissmann2013selecting}. The dependence captured by a pair-copula is characterized by a set of family-specific parameters governing the strength of the relationship between the corresponding variables. The probability that the two variables simultaneously take extremely high or low values can be  summarized by the copula's tail dependence coefficients, admitting values in $[0,1]$, with values close to zero indicating tail independence and those close to one indicating strong dependence~\cite{nelsen2006introduction}. A more detailed characterization of tail dependence for each copula family is provided in Appendix~\ref{app:families}. A key property of bivariate copulas is that conditional distributions can be expressed through the partial derivatives of the copula~\cite{schepsmeier2014derivatives}. For notational convenience, we denote these conditional distribution functions as h-functions, with:
$$
\h{1}{C}{u_1}{u_2}=\frac{\partial C(u_1,u_2)}{\partial u_2},
\qquad
\h{2}{C}{u_1}{u_2}=\frac{\partial C(u_1,u_2)}{\partial u_1}.
$$
These functions play a central role in PCCs, enabling the recursive construction of higher-dimensional dependence models starting from bivariate copulas.

\begin{figure}[!b]
    \centering
    \includegraphics[width=0.95\linewidth]{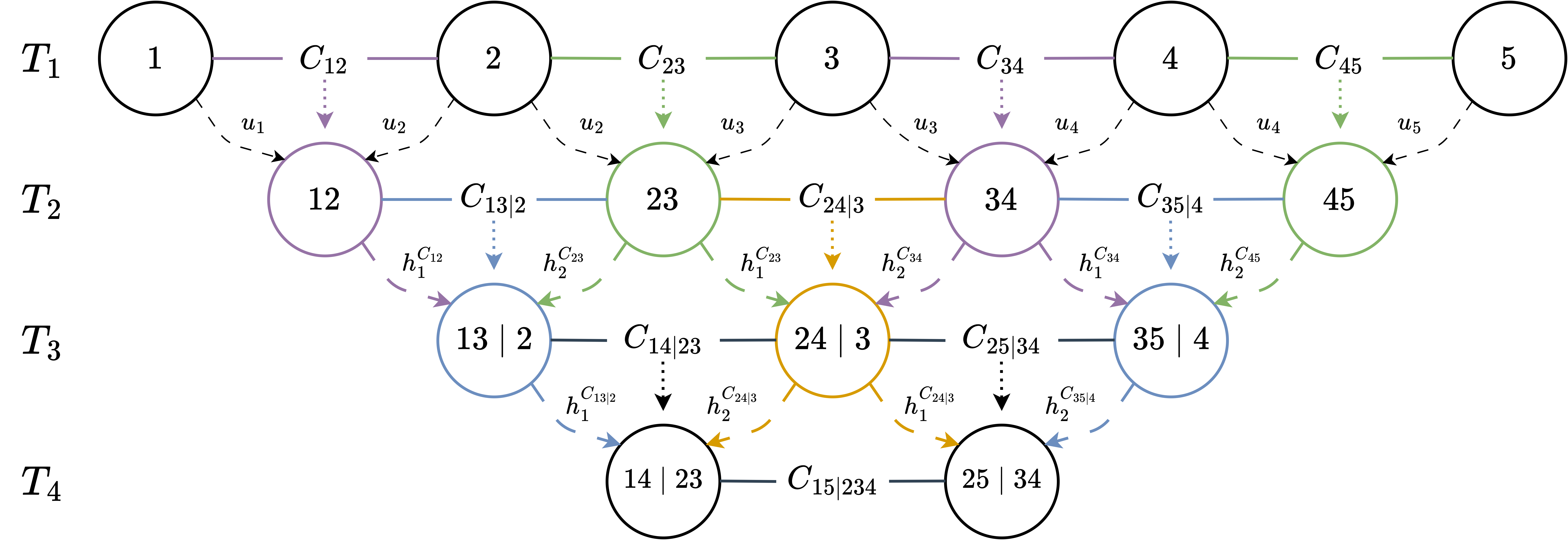}
    \caption{D-vine decomposition of $d=5$ variables. Each row is a tree $T_1$-$T_4$ and each edge a pair-copula. Nodes in $T_1$ are the modeled variables, while in subsequent trees they correspond to the edges of the previous tree. Dashed arrows represent the h-functions (or pseudo observations $u_i$ in $T_1$) that propagate data from one tree to the next.}
    \label{fig:dvine}
\end{figure}

A PCC in $d$ dimensions consists of $d(d-1)/2$ pair-copulas~\cite{aas2009pair}. A common PCC structure is the \emph{D-vine}, which arranges these pair-copulas as edges across $d-1$ linked trees (Fig.~\ref{fig:dvine}). The first tree contains unconditional pair-copulas between adjacent variables in a chosen ordering. Higher-level trees introduce conditional pair-copulas between variables that are progressively farther apart, conditioned on the variables lying between them. The required conditional distributions are computed recursively through h-functions~\cite{hobaek2013parameter}.

This construction yields the factorization of a $d$-dimensional density as:
\begin{equation}
\label{eq:densitydecompd}
f(x_1,\dots,x_d) = \prod_{i=1}^d f_i(x_i) \cdot \prod_{\ell=1}^{d-1}\prod_{i=1}^{d-\ell} c_{i,i+\ell\mid D_{i,\ell}}(F_{i\mid D_{i,\ell}}, F_{i+\ell\mid D_{i,\ell}})
\end{equation}
where $D_{i,\ell}={i+1,\dots,i+\ell-1}$ denotes the set of intermediate variables conditioning the pair-copula $c_{i,i+\ell\mid D_{i,\ell}}$~\cite{czado2019analyzing}.
For example, when $d=3$:
$$
f(x_1,x_2,x_3)
=
f_1(x_1)f_2(x_2)f_3(x_3)\,
c_{12}(u_1,u_2)\,
c_{23}(u_2,u_3)\,
c_{13|2}(u_{1|2},u_{3|2}),
$$
where $u_{1|2}$ and $u_{3|2}$ are obtained via h-functions.

Fitting a D-vine requires selecting both a copula family and its parameters for each of the $d(d-1)/2$ pair-copulas~\cite{dissmann2013selecting}. Since the number of possible configurations grows combinatorially with the number of candidate families and variable orderings, exhaustive search quickly becomes unfeasible~\cite{chapon2023imputation}. Moreover, pair-copulas cannot be optimized independently: the conditional distributions used in higher-level trees depend on the h-functions of lower-level trees, causing early modeling decisions to propagate throughout the vine~\cite{hobaek2013parameter}.

\subsection{Conformal Predictions}\label{subsec:cp}
Conformal prediction provides distribution-free uncertainty quantification by associating each test sample $\mathbf{x}$ with a prediction region rather than a single point prediction~\cite{angelopoulos2023conformal}.  Given a trained classifier, a separate calibration set $\mathcal{D}_{\mathrm{cal}}$ is used to compute non-conformity scores $\ncf_i$ for calibration examples. Given a significance level $\alpha$, for a candidate label $y$, the test point $(x,y)$ is assigned a non-conformity score $\ncf(x,y)$, which is compared against the threshold computed on the calibration scores~\cite{fontana2023conformal}: 
$$
\tau_\alpha =\frac{\lceil(\mid\mathcal{D}_{\mathrm{cal}}\mid+1)(1-\alpha)\rceil}{\mid\mathcal{D}_{\mathrm{cal}}\mid}\text{-th quantile of }\{\ncf_i:i\in\mathcal{D}_{\mathrm{cal}}\}.
$$
The prediction region is then defined as: $
\Gamma_\alpha(x) = \{y\in\{0,1\}: \ncf(x,y)\leq\tau_\alpha\}.
$

In binary classification, $\Gamma_\alpha(x)$ may contain one label (confident prediction), both labels (uncertainty), or, more rarely, no label. \emph{Mondrian} conformal prediction computes calibration scores separately for each class, yielding class-conditional coverage guarantees~\cite{vovk2005algorithmic}. Under the assumption of exchangeability between calibration and test observations belonging to the same class:
$$
\mathbb{P}_{(x,y)}\bigl(y\in\Gamma_\alpha(x)\bigr)
\geq
1-\alpha,
\qquad y\in\{0,1\},
$$
meaning that the prediction set contains the true label with probability at least $1-\alpha$ for each class individually. These guarantees hold in finite samples and do not rely on assumptions about the underlying data distribution~\cite{fontana2023conformal}.
\section{Methodology}
\label{sec:methodology}

This section presents our differentiable beam-search procedure for fitting D-vine copulas. We address two key challenges of D-vine model selection: the combinatorial growth of possible family assignments and the error propagation induced by greedy sequential fitting. The method combines gradient-based maximum likelihood estimation of pair-copula parameters with a beam-search strategy that maintains multiple candidate D-vine configurations during the fitting process.

\subsection{Fitting D-Vine Copulas}
\label{subsec:Fitting}

We assume access to a dataset of continuous variables suitable for copula modeling, where each observation is labeled as either ordinary or anomalous. The data are partitioned into three disjoint subsets: training, calibration and test. We denote with $\mathbf{X}_{r}^{y}$ the subset corresponding to split $r \in \{\mathrm{tr}, \mathrm{ca}, \mathrm{te}\}$ and label $y \subset \{0,1\}$, where $0$ denotes ordinary observations and $1$ anomalous ones. Thus, $\mathbf{X}_{\mathrm{tr}}^{(0,1)}$ are used for model fitting, $\mathbf{X}_{\mathrm{ca}}^{(0,1)}$ for calibration, and $\mathbf{X}_{\mathrm{te}}^{(0,1)}$ for evaluation. 

Given a dataset $\mathbf{X} \equiv \mathbf{X}^0_\mathrm{tr} \in \mathbb{R}^{n\times d}$ of $n$ \emph{ordinary} observations of dimension $d$, we first transform each variable to the
unit interval. For each variable $\mathbf{X}_{\cdot j}$, we estimate its marginal CDF $\hat F_j$ using Gaussian kernel density estimation
(KDE)~\cite{silverman2018density} and obtain pseudo-observations
$
\mathbf{U}_{ij} = \hat F_j(\mathbf{X}_{ij})$.
By the probability integral transform~\cite{yan2007enjoy}, the resulting matrix
$\mathbf{U} \in (0,1)^{n\times d}$ contains approximately uniform marginals and forms the input to the D-vine fitting procedure.

The objective is to fit a D-vine copula by selecting an appropriate pair-copula family for each edge and estimating the corresponding parameters.
At a high level, the procedure alternates between local fitting and global search. For every edge in the current tree, all candidate copula families are fitted independently by gradient-based maximum likelihood estimation. Rather than committing immediately to the best-fitting family, statistically indistinguishable alternatives are retained. These alternatives generate multiple extensions of the current partial D-vine configurations. A beam-search strategy then preserves only the most promising configurations according to their cumulative log-likelihood. Once all trees have been processed, the surviving complete D-vine configurations undergo a final joint parameter refinement.

\begin{figure}[!t]
    \centering
    \includegraphics[width=1\linewidth]{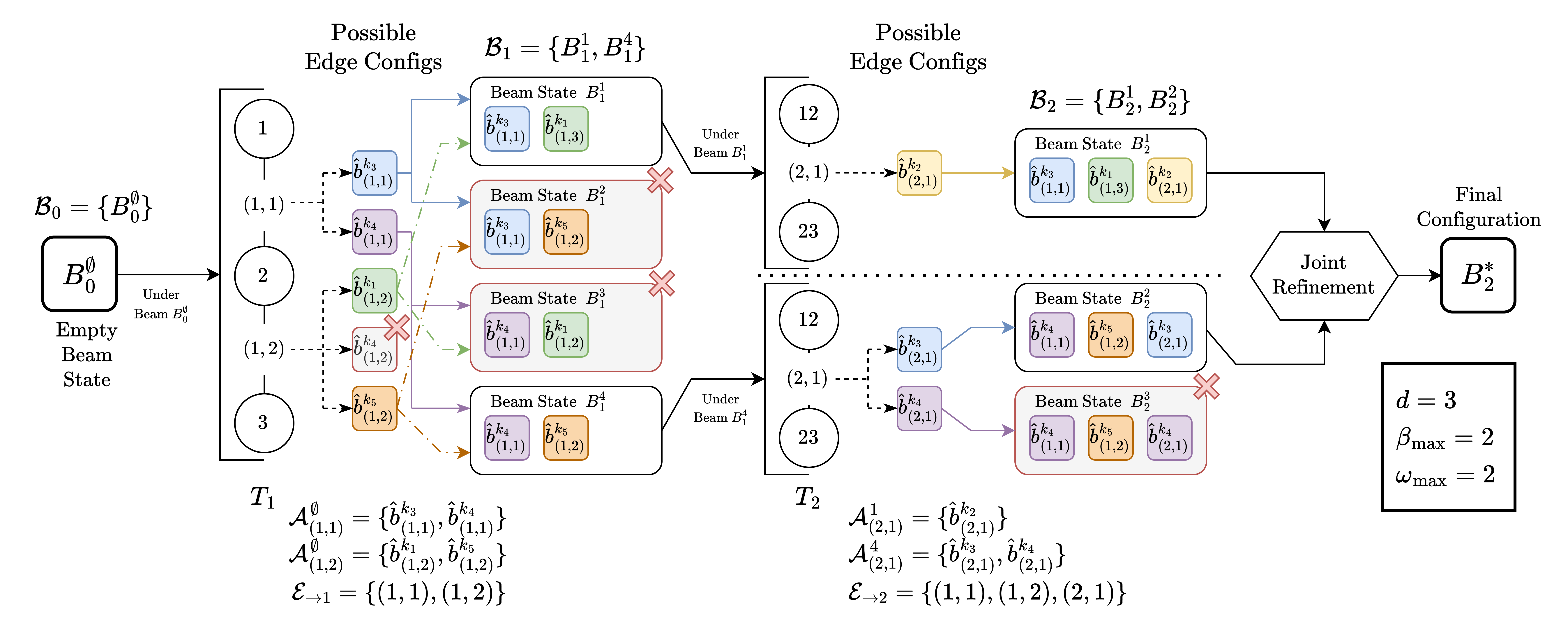}
    \caption{Graphical representation of the proposed D-vine fitting methodology for $d=3$ variables, with $\beta_\text{max}=2$ and $\omega_\text{max}=2$. Trees are processed sequentially from an empty beam state $B_0^\emptyset$. Each edge in $T_1$ is individually fitted, keeping at most the best $\omega_\text{max}$ configurations for each node. These are combined via their Cartesian product into candidate beam states $B_1^1,\dots,B_1^4$, of which the best $\beta_{\text{max}}$ form $\mathcal{B}_1$. The procedure repeats at $T_2$, with the best final beams undergoing joint refinement yielding the final configuration $B_2^*$. Crossed out states are discarded configurations.}
    \label{fig:dvine_fitting}
\end{figure}

We consider a candidate family set $
\mathcal{F}=\{F_1,\dots,F_K\}$.
For each pair-copula in the D-vine, the goal is to select a family $F_k\in\mathcal{F}$ and estimate its associated parameter vector $\theta_k$. The inputs to a generic pair-copula are denoted by $u,v\in [0,1]^n$. In the first tree, these correspond to pseudo-observations obtained directly from the data, whereas in higher-order trees, they are produced through the cascade of h-functions. A generic pair-copula is denoted by $C_{ij\mid D}$, where $i$ and $j$ identify the coupled variables and $D$ is the conditioning set.
To simplify notation, each pair-copula is uniquely identified by its edge position $(t,p)$ in the D-vine, where $t\in\{1,\dots,d-1\}$
is the tree index and $p\in\{1,\dots,d-t\}$
is the edge position within tree $T_t$. We use the shorthand notation
$c_{(t,p)}$, $
h_1^{(t,p)}$ and
$h_2^{(t,p)}$
for the corresponding copula density and h-functions associated with the underlying pair-copula $C_{ij\mid D}$ (see Fig.~\ref{fig:dvine_fitting}).

Since the D-vine is fitted sequentially across trees $T_1,\dots,T_{d-1}$, we define the set of edges processed up to tree $T_t$ :
$$
\mathcal{E}_{\to t}
:= \{(t',p')
\mid
t'=1,\dots,t,;
p'=1,\dots,d-t'\}.$$

A \emph{beam state} $B_t^i$ represents a partial D-vine configuration after processing tree $T_t$. It stores the selected family and parameter values for all edges in $\mathcal{E}_{\to t}$:
$$
B_t^i :=
\big\{
b_{(t',p')}
\mid
(t',p')\in\mathcal{E}_{\to t}
\big\},
$$
where
$b_{(t',p')}=
(F_k,\theta_k)$
is the selected family--parameter configuration for edge $(t',p')$. Note that every edge $(t',p')$ may be assigned to a different family $F_k$.
The set
$\mathcal{B}_t
=\{B_t^1,B_t^2,\dots\}$
contains all active beam states after processing tree $T_t$.
For edge $(t,p)$ and family--parameter configuration
$b_{(t,p)}=(F_k,\theta_k)$, for some $k\in\{1,\dots, K\}$, the \emph{pointwise log-likelihood} contribution of observation $m$ is:
\begin{equation}
\label{eq:pointLL}
\ell(b_{(t,p)};u_m,v_m) =
\log \big(c_{(t,p)}(u_m,v_m;\theta_k)\big),
\end{equation}
where $c_{(t,p)}(\cdot,\cdot;\theta_k)$ is the copula density of family $F_k$.
The \emph{edge-specific log-likelihood} is then computed as:
\begin{equation}\label{eq:edge_lkh}
\ell(b_{(t,p)}; \mathbf{U})
=\sum_{m=1}^{n}
\ell(b_{(t,p)};u_m,v_m).
\end{equation}
The \emph{score} of a beam state is defined as the cumulative log-likelihood of all edges it contains:
\begin{equation}
\ell(B_t^i; \mathbf{U})
=\sum_{(t',p')\in\mathcal{E}_{\to t}}
\ell\left(b_{(t',p')}; \mathbf{U}\right).
\end{equation}

The search breadth is controlled by two user-defined hyperparameters: the maximum number of beam states retained after each tree $\beta_{\max}$, and the maximum number of candidate configurations retained for each edge $\omega_{\max}$.

The algorithm is initialized with a single empty beam state $B_0^\emptyset$, with $\mathcal{B}_{0}=\{B_0^\emptyset\}$.
Fitting proceeds sequentially over the trees $T_1,\dots,T_{d-1}$. At iteration $t$, each active beam state
$B_{t-1}^i\in\mathcal{B}_{t-1}$
is expanded independently.
For every edge $(t,p)$ in the current tree and for every candidate family $F_k\in\mathcal{F}$, parameters are estimated through gradient-based maximization of the edge-specific log-likelihood defined in~\eqref{eq:edge_lkh}:
\begin{equation}
  \hat{\theta}_k
=
\arg\max_{\theta_k}\
\ell(b_{(t,p)};\mathbf{U}).  
\end{equation}
The fitted configuration
$
\hat{b}_{(t,p)}
=(F_k,\hat{\theta}_k)
$
is associated with edge log-likelihood
$
\ell(\hat{b}_{(t,p)};\mathbf{U}).
$
After fitting all the $K$ candidate families, the configuration with the highest edge log-likelihood is identified as the leading candidate. However, selecting only the top-scoring family may discard alternatives that are statistically indistinguishable and could lead to better overall D-vine configurations.

To detect such cases, we compare the leading candidate against the other configurations using either the Vuong test~\cite{vuong1989likelihood} or the Clarke test~\cite{clarke2007simple}. Both tests assess whether differences in likelihood are statistically significant. If the leading candidate is significantly better than every alternative, the edge commits to it. Otherwise, all non-rejected alternatives are retained. Since the Clarke test is generally more conservative, it typically produces larger candidate sets.

To limit combinatorial growth, at most $\omega_{\max}$ candidate configurations are retained for each edge, prioritizing those with the highest edge log-likelihood. This produces the candidate set$$
\mathcal{A}_{(t,p)}^i
=\{
\hat{b}_{(t,p)}^1,\dots,
\hat{b}_{(t,p)}^\omega
\},
\qquad
\omega\leq \omega_{\max},
$$
for edge $(t,p)$ under parent beam state $B_{t-1}^i$.
Once candidate sets have been obtained for all edges in the current tree, they are combined through the Cartesian product
$
\mathcal{A}_t^i
=
\mathcal{A}_{(t,1)}^i
\times
\dots
\times
\mathcal{A}_{(t,d-t)}^i.
$
Each element
$
a\in\mathcal{A}_t^i
$
defines a possible extension of the parent beam state $B_{t-1}^i$. The resulting beam score is obtained by adding the cumulative score of the parent beam to the edge log-likelihoods contributed by the configurations in $a$. Repeating this process for every active beam state in $\mathcal{B}_{t-1}$ generates a pool of candidate beam states for tree $T_t$. From this pool, only the $\beta_{\max}$ states with the highest beam score are retained, forming the beam set $\mathcal{B}_t$ passed to the next iteration.

After processing all trees, the final beam set $\mathcal{B}_{d-1}$ contains complete D-vine configurations. Although these configurations achieve high cumulative log-likelihood, pair-copula parameters have been estimated locally while treating the outputs of previously fitted copulas as fixed. Since the inputs of a pair-copula in tree $T_t$ are generated through the h-function cascade of preceding trees, parameter updates can affect all downstream pair-copulas.

To account for these dependencies, each beam state in $\mathcal{B}_{d-1}$ undergoes a final \emph{joint refinement} phase. Keeping the selected copula families fixed, all pair-copula parameters are optimized simultaneously by maximizing the full D-vine log-likelihood. Optimization is initialized from the locally fitted parameters and performed end-to-end, allowing gradients to propagate through the complete h-function cascade and explicitly accounting for cross-tree parameter interactions.

The beam state with the highest refined log-likelihood is selected as the final fitted D-vine configuration:
\begin{equation}
   B_{d-1}^{*}
=
\arg\max_{B\in\mathcal{B}_{d-1}}
\ell(B;\mathbf{U}). 
\end{equation}
Fig.~\ref{fig:dvine_fitting} illustrates the full fitting procedure on an example with $d=3$ variables.

\paragraph{Variable Ordering.}
The ordering of variables determines the unconditional pair-copulas in the first tree $T_1$ and therefore strongly influences the quality of the resulting D-vine. Following the approach of~\cite{dissmann2013selecting}, we construct the ordering by placing strongly dependent variables adjacent in the sequence, quantifying dependence between each pair of variables through the empirical Kendall's Tau~\cite{nelsen2006introduction}. The final ordering is the path through the $d$ variables maximizing the sum of absolute pairwise Kendall's Tau over adjacent pairs.

\paragraph{Anomaly-aware fitting.}
The D-vine is fitted on $\mathbf{X}_{\mathrm{tr}}^{0}$ using the procedure described above. When anomalous training observations $\mathbf{X}_{\mathrm{tr}}^{1}$ are available, we augment the fitting objective with a penalization term that discourages high likelihood values on anomalous samples. Specifically, Eq.~\ref{eq:edge_lkh} becomes
\begin{equation}
\label{eq:penEdgeLL}
\ell^{\mathrm{pen}}(b_{(t,p)};\mathbf{U})
=
\ell(b_{(t,p)};\mathbf{U}^{0})
-
\lambda\,\ell(b_{(t,p)};\mathbf{U}^{1}),
\end{equation}
where $\mathbf{U}^{0}$ and $\mathbf{U}^{1}$ are the pseudo-observations derived from $\mathbf{X}_{\mathrm{tr}}^{0}$ and $\mathbf{X}_{\mathrm{tr}}^{1}$, respectively, and $\lambda \geq 0$ controls the penalization strength. This encourages the fitted D-vine to capture dependence structures characteristic of ordinary observations while reducing its fit to anomalous ones.

\subsection{Anomaly Detection}
\label{subsec:AD}

We now describe how the fitted D-vine is used to perform anomaly detection. The proposed procedure leverages the hierarchical structure of the vine decomposition to produce both a global anomaly score and per-edge scores that facilitate the identification of the variable relationships responsible for a detected anomaly.

\paragraph{Edge-level anomaly scores.}

The anomaly detection criterion is based on the negative log-copula density associated with each pair-copula in the fitted D-vine under configuration $B_{d-1}^*$. Given an observation $\mathbf{u}_m$, we compute the edge-wise scores $-\ell(b^*_{(t,p)};\mathbf{u}_m)$. These values are large when the observed dependence pattern deviates from that expected under the fitted pair-copula and small when the two are consistent.
To establish a baseline under ordinary conditions, we evaluate these scores on the ordinary training set $\trainClean$. Since pair-copulas belonging to different families and parameterizations may produce likelihood values on substantially different scales, the resulting scores are not directly comparable across edges. We therefore standardize each edge independently using the median and median absolute deviation (MAD) of its training scores:
\begin{align}
\text{med}_{(t,p)} &= \underset{m \in \trainClean}{\text{median}} \big( -\ell(b^*_{(t,p)};\mathbf{u}_m)\big), \\
\text{MAD}_{(t,p)} &= \underset{m \in \trainClean}{\text{median}} \big| {-\ell(b^*_{(t,p)};\mathbf{u}_m)} - \text{med}_{(t,p)} \big|.
\end{align}
The median and MAD are preferred over the mean and standard deviation because of their robustness to outliers. For a generic observation $\mathbf{u}_m$, the standardized score at edge $(t,p)$ is defined as
\begin{equation}
\mathtt{score}\big((t,p), \mathbf{u}_m\big)
=\frac{-\ell(b^*_{(t,p)};\mathbf{u}_m)-\text{med}_{(t,p)}}
{\text{MAD}_{(t,p)}},
\end{equation}
which can be interpreted as the number of MAD units by which the edge score deviates from the ordinary baseline.
Collecting these values over all edges yields the score vector
$$
\mathbf{s}_m
=\Big\{
\mathtt{score}\big((t,p),\mathbf{u}_m\big)
:
(t,p)\in\mathcal{E}_{\to d-1}
\Big\},
$$
which summarizes the contribution of each pair-copula to the anomaly assessment and forms the basis for both global detection and local interpretation.

\paragraph{Global anomaly score.}

While various anomaly detection strategies can be applied to $\mathbf{s}_m$, we employ a simple aggregation rule and define a scalar anomaly score $S_m$ as the average of the $\kappa$ largest edge values in $\mathbf{s}_m$:
\begin{equation}
\label{eq:aggScore}
S_m
=
\frac{1}{\kappa}
\sum_{(t,p)\in\mathcal{T}_\kappa^{(m)}}
\mathtt{score}\big((t,p),\mathbf{u}_m\big),
\end{equation}
where $\mathcal{T}_\kappa^{(m)}$ denotes the set of the $\kappa$ highest-scoring edges for observation $m$ and $\kappa\in\{1,\dots,d(d-1)/2\}$ is a user-configurable parameter.

\paragraph{Conformal anomaly detection.}

The aggregate score $S_m$ provides a scalar measure of abnormality. We next convert this score into a statistically calibrated anomaly detector using split conformal prediction (CP)~\cite{angelopoulos2023conformal}, introduced in Sect.~\ref{subsec:cp}.

We formulate anomaly detection as a binary classification problem, assigning label $0$ to ordinary observations and label $1$ to anomalous observations. Given a test observation $\mathbf{x}_m$ with unknown label $y_m$, let $S_m$ denote its aggregate score computed from the corresponding pseudo-observation $\mathbf{u}_m$. We assess the compatibility of $S_m$ with each class through class-conditional non-conformity functions
$$
\ncf(S_m;0) = S_m - \mathtt{d_S},
\qquad
\ncf(S_m;1) = \mathtt{d_S} - S_m,
$$
where $\mathtt{d_S}$ is a threshold separating ordinary and anomalous scores. In our experiments, $\mathtt{d_S}$ is chosen as the 95th percentile of the aggregate scores computed over $\trainClean$. The non-conformity score is small when the observation is consistent with the corresponding class and large otherwise.

Given a desired confidence level $\alpha$, we adopt the Mondrian conformal framework~\cite{vovk2005algorithmic} and calibrate the two classes separately by applying $\ncf$ to the labeled observations in $\calibClean$ and $\calibAnom$, respectively. The thresholds $\tau_\alpha^0$ and $\tau_\alpha^1$ are then defined as the
$\lceil (n_0+1)(1-\alpha)\rceil/n_0$ and $
\lceil (n_1+1)(1-\alpha)\rceil/n_1
$
empirical quantiles of the corresponding non-conformity scores, where $n_0=|\calibClean|$ and $n_1=|\calibAnom|$.
The \emph{prediction region} associated with $\mathbf{x}_m$ is:
\begin{equation}
\Gamma_\alpha(\mathbf{x}_m)
=
\big\{
y\in\{0,1\}
:
\ncf(S_m,y)\le\tau_\alpha^y
\big\}.
\end{equation}
An observation receives label $0$ when $\ncf(S_m,0)\le\tau_\alpha^0$ and label $1$ when $\ncf(S_m,1)\le\tau_\alpha^1$. Consequently, the prediction region can be $\{0\}$, $\{1\}$, $\{0,1\}$, or $\emptyset$, corresponding respectively to a unique ordinary prediction, a unique anomalous prediction, ambiguity between the two classes, or incompatibility with both.

Under the assumption of exchangeability between calibration and test observations belonging to the same class, Mondrian conformal prediction \emph{guarantees class-conditional coverage}:
$$
\mathbb{P}_{(\mathbf{x}_m,y_m)}
\bigl(
y_m\in\Gamma_\alpha(\mathbf{x}_m)
\bigr)
\ge
1-\alpha,
$$
meaning that the prediction region contains the true label with probability at least $1-\alpha$ for each class~\cite{vovk2005algorithmic}. This distinction is particularly important in anomaly detection, where the anomalous class is often heavily underrepresented. Standard conformal prediction guarantees only marginal coverage over the entire population, which may be dominated by ordinary observations and therefore provides substantially weaker guarantees for anomalies. By calibrating each class separately, Mondrian conformal prediction ensures the desired coverage level for both ordinary and anomalous observations.

\paragraph{Interpretability.}

The standardized edge scores naturally provide an interpretable explanation of detected anomalies. Each pair-copula $C_{ij\mid D}$ models the dependence between variables $i$ and $j$ conditional on a set $D$. Consequently, a large score at the associated edge indicates that the anomaly is associated with an unexpected dependence pattern in that specific relationship. Rather than merely detecting the presence of an anomaly, the proposed framework identifies which variable interactions contribute most strongly to the anomalous behavior.

\section{Experiments}\label{sec:experiments}

\paragraph{Pair-copula implementations.}
To support gradient-based fitting, we developed a fully differentiable PyTorch implementation of the Gaussian and Student-t elliptical copulas, together with the Clayton, Frank, Gumbel, and Joe Archimedean families. Copula densities and h-functions are implemented in a differentiable and numerically stable manner, with real-valued parameters mapped to their valid domains through scaled sigmoid transformations. The supported families span a broad range of dependence structures; parameter bounds, tail-dependence properties and closed-form expressions with illustrative examples are provided in Appendix~\ref{app:families}. Numerical stability is further ensured through log-space density evaluation, input clamping away from 0 and 1, and masking of non-finite likelihood contributions.

\paragraph{Experimental setup.} We evaluate the proposed methodology on two datasets: the Wilt dataset from the ADBench benchmark~\cite{han2022adbench} and a proprietary real-world dataset of operational measurements from a sewerage network in northern Italy. Both datasets are well suited to copula-based modeling, as they consist exclusively of continuous variables, contain a sufficient number of observations and labeled anomalies to support both training and evaluation, and involve a relatively small number of features, resulting in compact D-vine decompositions.

Unless otherwise stated, we report results for the penalized fitting procedure introduced in Eq.~\ref{eq:penEdgeLL} and the corresponding anomaly detection framework. Comparisons with the non-penalized variant are discussed where relevant, while complete results are provided in Appendix~\ref{app:results}.

For both datasets, the candidate family set comprises the Gaussian, Student-t, Clayton, Joe, Frank, and Gumbel copulas.
During fitting, each pair-copula is optimized for 250 epochs in the initial edge-wise stage, followed by 200 epochs of joint refinement for each surviving beam state in $\mathcal{B}_{d-1}$. The Clarke test is used to evaluate competing edge configurations with the maximum branching factor set to $\omega_{\max}=4$, while the beam width is limited to $\beta_{\max}=8$. Unless otherwise stated, the anomaly penalization coefficient in~\eqref{eq:penEdgeLL} is fixed at $\lambda=0.1$.

\begin{figure}[!t]
    \centering
    \includegraphics[width=0.9\linewidth]{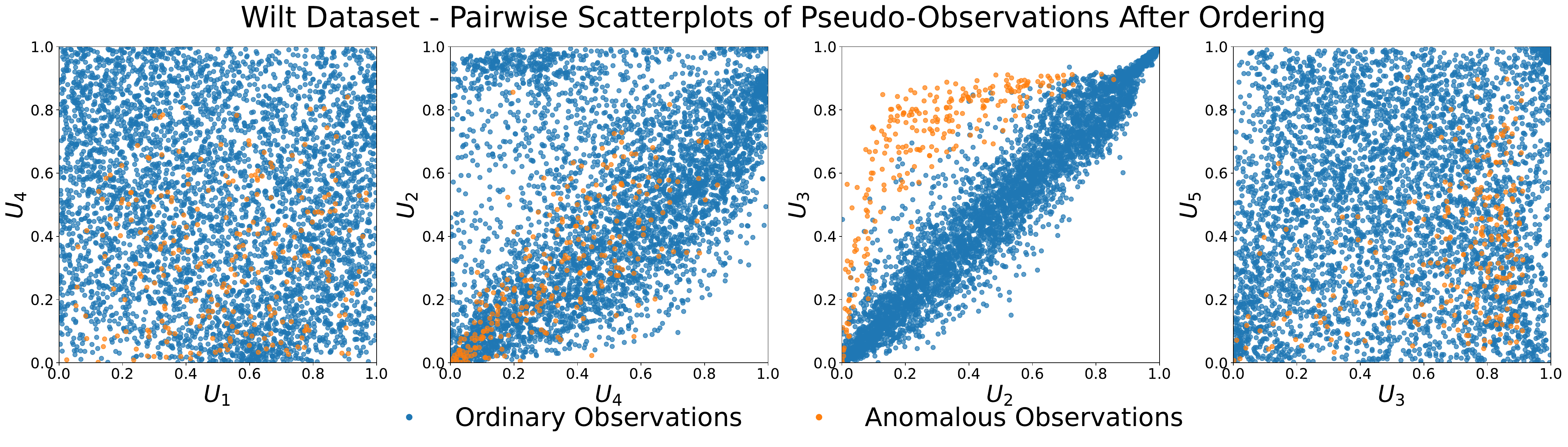}
    \caption{\textbf{Wilt:} Pairwise scatterplots of the ordered pseudo-observations. Variables pairs plotted together form the inputs to the pair-copulas at the first tree $T_1$.}
    \vspace{-.25cm}
    \label{fig:WiltPSO}
\end{figure}
\begin{figure}[!b]
\vspace{-.25cm}
\centering
\includegraphics[width=1\linewidth]{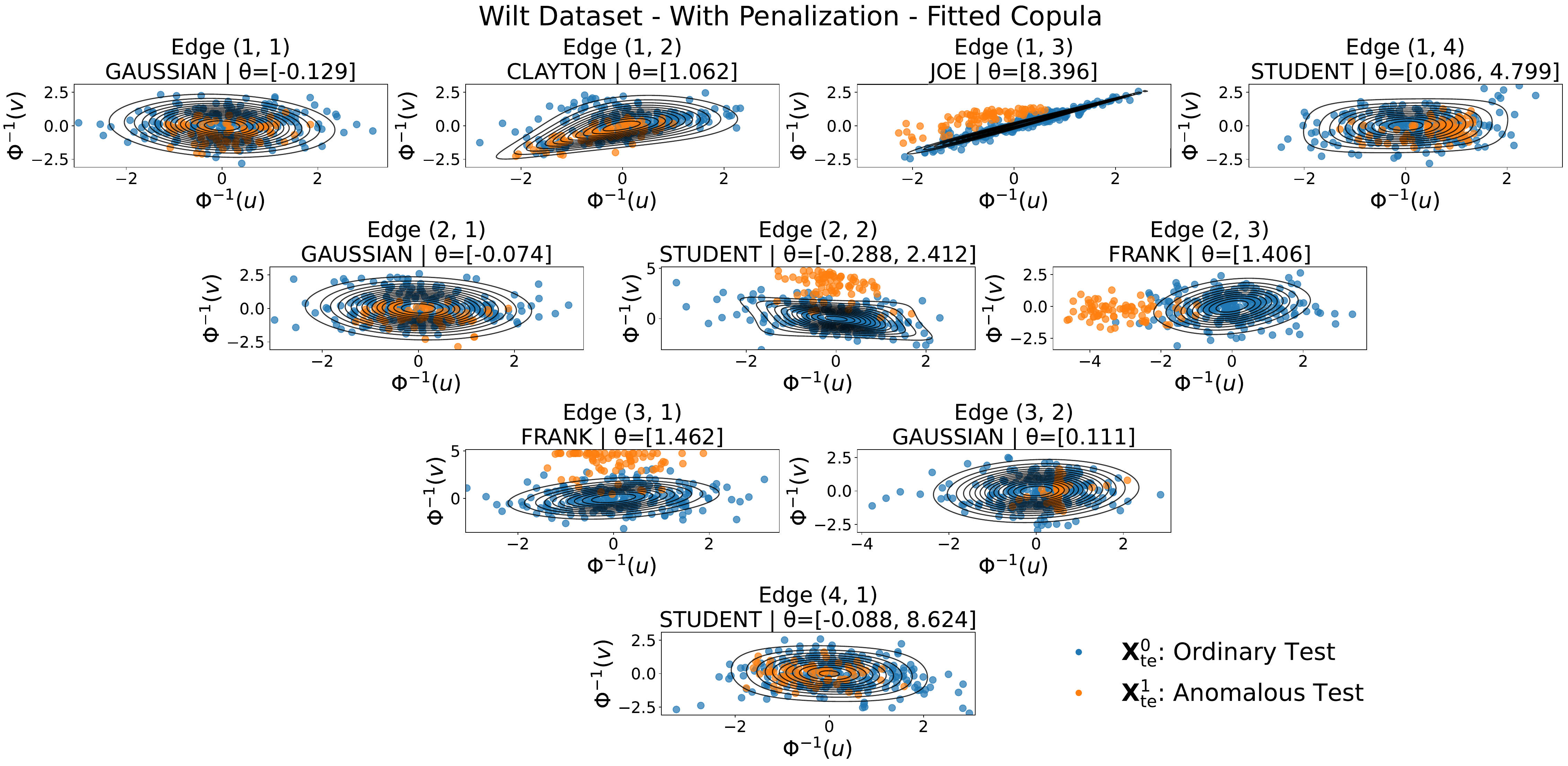}
\caption{\textbf{Wilt:} Fitted pair-copulas of the D-vine, shown as joint densities. Each panel can be read in isolation, as the dependence accumulated in earlier trees is embedded in the inputs of each panel through the h-function cascade.}
\label{fig:WiltedFittedYesPen}
\end{figure}
\subsection{Wilt Dataset Results}
The Wilt dataset is derived from a remote-sensing task in which satellite image segments of forested areas are classified as either diseased trees or other land cover, with the diseased-tree segments treated as anomalies. It is comprised of roughly 4500 ordinary and 250 anomalous observations across five spectral features. After applying the estimated marginal CDFs and variable ordering, the resulting pseudo-observations reveal two different dependence patterns for ordinary and anomalous observations, as can be seen in the third panel of Fig.~\ref{fig:WiltPSO},  making the dataset an ideal candidate for our copula-based approach.

Fig.~\ref{fig:WiltedFittedYesPen} shows the D-vine configuration estimated by the proposed fitting procedure. Each panel visualizes the joint density induced by the fitted pair-copula combined with standard Gaussian marginals. Since the true marginal distributions of conditional variables are generally unavailable beyond the first tree, Gaussian marginals are adopted solely for visualization purposes. This choice does not affect the interpretation of the dependence structure, as copulas are invariant to the marginal distributions~\cite{nelsen2006introduction}.

\begin{figure}[!t]
    \centering
    \includegraphics[width=0.95\linewidth]{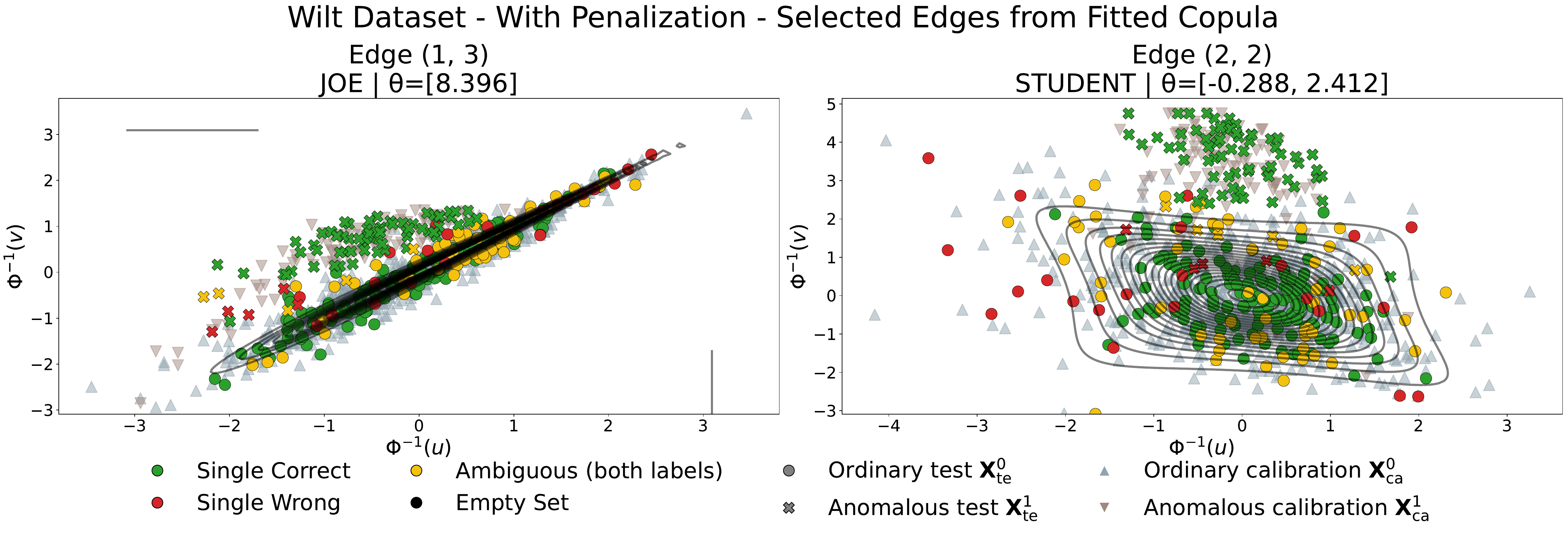}
    \caption{\textbf{Wilt:}  Anomaly detection results on the most relevant edges; color indicates prediction correctness, shape distinguishes ordinary from anomalous observations.}
    \label{fig:wiltedAD}
    \vspace{-.3cm}
\end{figure}

We notice that the penalization does not affect all edges uniformly as most retain similar family parameter configurations as in the non penalized fit (Appendix~\ref{app:results}), with its effects being concentrated on edges $(1,3)$ and $(2,2)$. Under penalization, edge $(1,3)$ switches from a Gumbel to a Joe copula, with the latter exhibiting stronger upper tail dependence. Edge $(2,2)$ maintains a Student-t copula but the value of the second parameter, the degrees of freedom, drops substantially, indicating heavier tails and a departure from Gaussian behavior. As can be seen in the third panel of Fig.~\ref{fig:WiltPSO} and in Fig.~\ref{fig:WiltedFittedYesPen}, $(1,3)$ is the edge in $T_1$ with the clearest separation between ordinary and anomalous observations and, as the inputs of $(2,2)$ are derived in part from $(1,3)$ through the h-function cascade, this separation propagates to the second tree.
\begin{figure}[!b]
    \centering
    \includegraphics[width=1\linewidth]{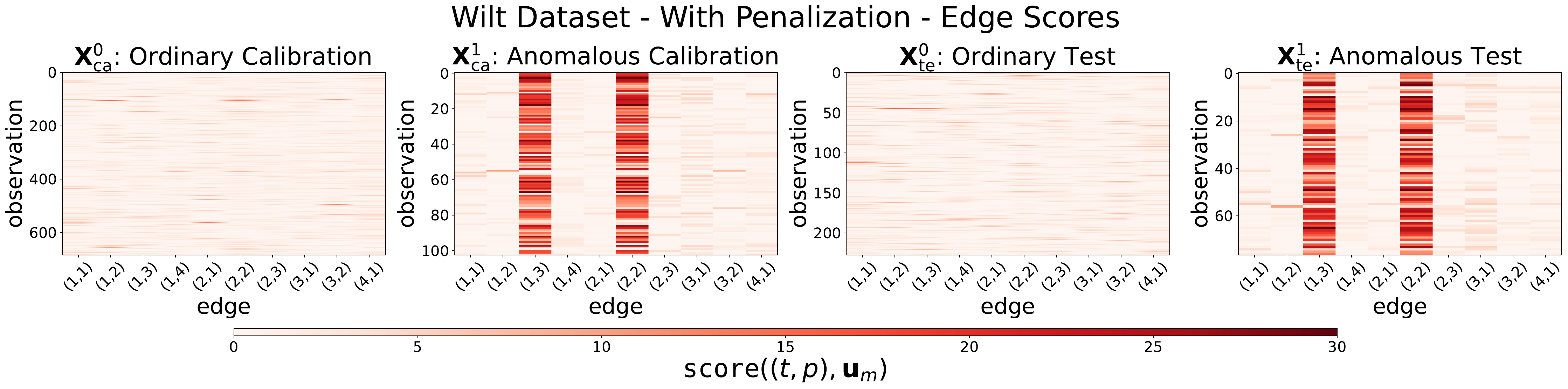}
    \caption{\textbf{Wilt:} Standardized edge scores $\mathtt{score}\big((t,p), \mathbf{u}_m\big)$ for the Wilt dataset across the calibration and test sets, for fitting with penalization.}
    \label{fig:wiltedAttribution}
\end{figure}
The relevance of these edges in identifying potential anomalies is also visible in Fig.~\ref{fig:wiltedAttribution}, which shows the standardized scores 
$\mathbf{s}_m$ for each edge in the decomposition and for each observation $m$ in $\calibClean$, $\calibAnom$, $\testClean$ and $\testAnom$. Edges $(1,3)$ and $(2,2)$ present significantly higher scores for observations belonging to the anomalous sets $\calibAnom$ and $\testAnom$ compared to those belonging to their ordinary counterparts, confirming that these edges model the variable relationships through which anomalies can be detected. Lastly, it can be noted that the calibration and test splits exhibit consistent patterns within each class, indicating that the standardized edge scores generalize from the calibration set to unseen observations. For this dataset, we compute the aggregate anomaly score following Eq.~\ref{eq:aggScore} with $\kappa=2$. After setting the significance level $\alpha = 0.1$, we calibrate the two class-conditional thresholds $\tau^{0}_{\alpha}$ and $\tau^{1}_{\alpha}$ on $\calibClean$ and $\calibAnom$ as per Sect.~\ref{subsec:AD} and assign each observation in $\testClean$ and $\testAnom$ a predictive region $\Gamma_{\alpha}(\mathbf{x}_m)$. Table~\ref{tab:wilt-cp} reports the class conditional conformal prediction results on the Wilt dataset for $\alpha=0.1$, with both the ordinary and anomalous classes reaching the target coverage level of $0.9$. 
Most observations receive a single correct label, none are assigned an empty set, and only a minority receive an ambiguous prediction. Fig.~\ref{fig:wiltedAD} provides a graphical view of the same result over the two most discriminative edges, $(1,3)$ and $(2,2)$. Correct single-label assignments concentrate where the two classes are well separated, while ambiguous or incorrect assignments tend to occur in the intermediate region where ordinary and anomalous observations overlap.

\begin{figure}[!b]
\vspace{-.3cm}
    \centering
    \includegraphics[width=\linewidth]{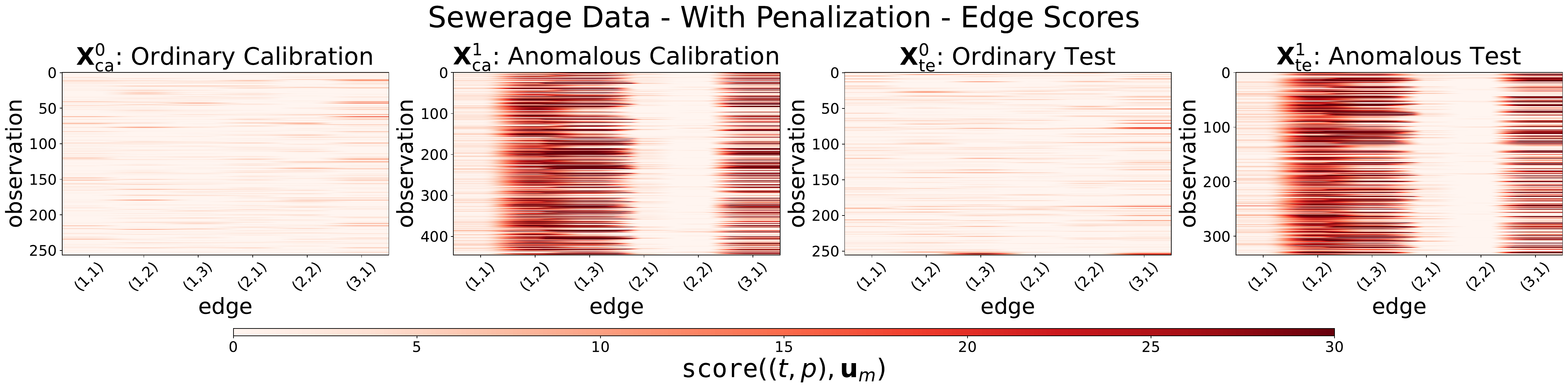}
    \vspace{-.5cm}
        \caption{\textbf{Sewerage:} Standardized edge scores $\mathtt{score}\big((t,p), \mathbf{u}_m\big)$ for the Sewerage dataset across the calibration and test sets, for fitting with penalization.}
    \label{fig:brianzaAttribution}
\end{figure}
\vspace{-.05cm}
\subsection{Sewerage Dataset Results}
Our second dataset is a multivariate time-series of real-world operational measurements from a sewerage network in northern Italy, containing four months of hourly level recordings from four sensors placed at neighbouring points within the same network, referred to as Sensors 1-4. Each recording is treated as an individual observation. Under ordinary dry-weather conditions, all four sensors follow the same regular daily pattern, and rainfall events produce simultaneous spikes in level across all four. During the observation period, Sensor 4 enters an anomalous state in which its readings, while remaining within an ordinary operational range, no longer follow the behavior shared by the rest of the network.

\begin{figure}[!t]
    \centering
    \includegraphics[width=0.95\linewidth]{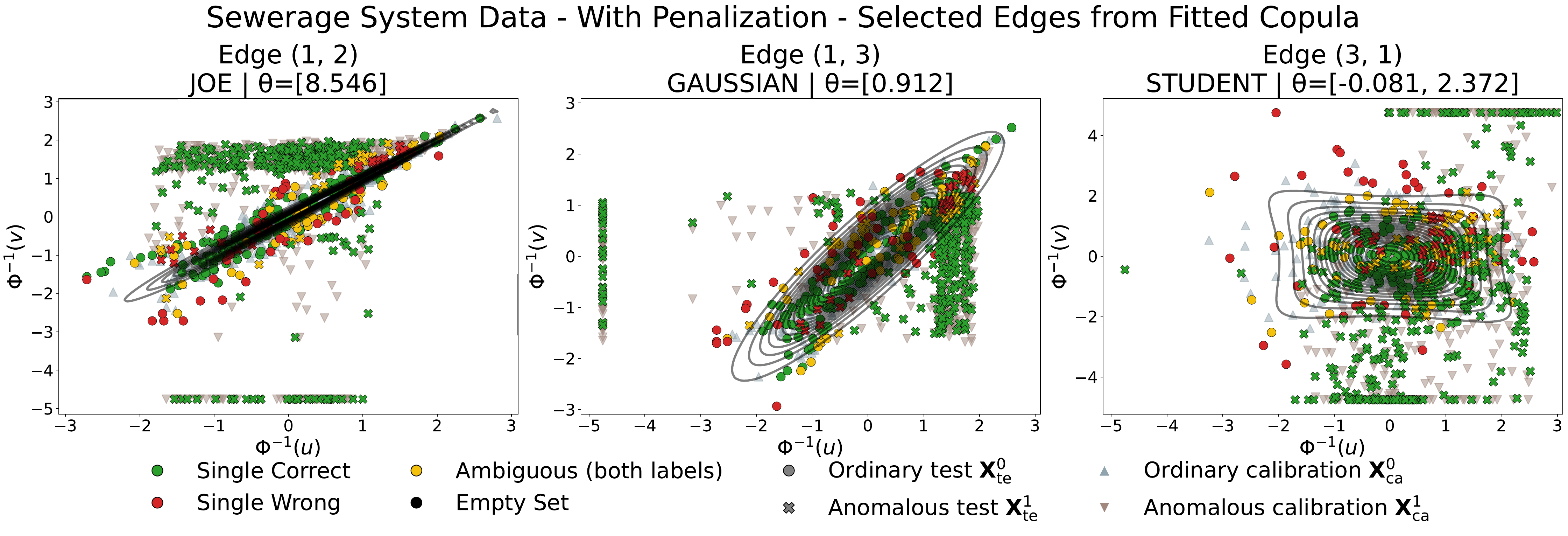}
    \vspace{-.25cm}
    \caption{\textbf{Sewerage:} Anomaly detection results on the most relevant edges; color indicates prediction correctness, shape distinguishes ordinary from anomalous observations.}
    \label{fig:brianzaSelectedEdges}
    \vspace{-.3cm}
\end{figure}

Following the same procedure as for the Wilt dataset, we fit a D-vine for the four sensors and report the resulting standardized edge scores $\mathbf{s}_m$ 
in Fig.~\ref{fig:brianzaAttribution}. Edges $(1,2)$, $(1,3)$ and $(3,1)$ clearly discriminate between ordinary and anomalous observations, with significantly higher values on the anomalous sets. Notably, after accounting for variable ordering, $(1,2)$ and $(1,3)$ are the pair-copulas in $T_1$ taking as input Sensor 4, demonstrating the method's ability to identify the cause of the deviation from expected behavior. Lastly, we compute the aggregate score of Eq.~\ref{eq:aggScore} with $\kappa=3$ and apply the same conformal procedure at $\alpha=0.1$, obtaining the predictive regions $\Gamma_\alpha$ for all observations in $\testClean$ and $\testAnom$ whose results are reported in Table~\ref{tab:wilt-cp} and Fig.~\ref{fig:brianzaSelectedEdges}. Both classes reach the desired coverage level and, as for the Wilt dataset, few observations are misclassified or assigned an ambiguous prediction, with none receiving an empty set. Correct single label assignments concentrate where the two classes are clearly separated. 

\newcolumntype{Y}{>{\centering\arraybackslash}X}
\begin{table}[!t]
\centering
\renewcommand{\arraystretch}{1.2}
\begin{tabularx}{\textwidth}{c|YYYYYYY}
\textbf{Dataset} &
\textbf{True Class} &
$\mathbf{n}$ &
\textbf{Coverage} &
\textbf{Single Correct} &
\textbf{Single Wrong} &
\textbf{Both Labels} &
\textbf{Empty Set} \\ \hline
\multirow{2}{*}{\begin{tabular}[c]{@{}c@{}}Wilt\\ Dataset\end{tabular}}
& Ordinary  & 228  & 0.904 & 151 & 22 & 55 & 0 \\ \cline{2-8}
& Anomalous &  77 &  0.935 & 67 & 5 & 5 & 0\\ \hline
\multirow{2}{*}{\begin{tabular}[c]{@{}c@{}}Sewerage\\ Data\end{tabular}}
& Ordinary  & 256 & 0.918 & 191 & 21 & 44 & 0 \\ \cline{2-8}
& Anomalous & 335 & 0.904 & 273 & 32 & 30 & 0 \\ \hline
\end{tabularx}
\caption{Class-conditional conformal prediction results on both datasets at $\alpha = 0.1$. Coverage is the fraction of observations whose prediction region contains the true label. }
\vspace{-.5cm}
\label{tab:wilt-cp}
\end{table}

\vspace{-.2cm}
\section{Conclusion}
\vspace{-.2cm}
We presented a novel framework for localized anomaly detection based on differentiable D-vine copulas. At its core is a beam-search fitting procedure that maintains multiple candidate D-vine configurations throughout the construction process, avoiding the myopic decisions of standard greedy methods. This is enabled by a fully differentiable PyTorch implementation of several pair-copula families, allowing gradient-based estimation and joint refinement of full D-vine configurations. Building on the fitted D-vine, we introduced an anomaly detection approach that exploits the hierarchical pair-copula decomposition to produce global anomaly scores and localized explanations. Class-conditional statistical guarantees are provided via Mondrian conformal prediction, enabling reliable uncertainty quantification even in unbalanced anomaly detection settings.

Experiments on both a public benchmark dataset and a real-world sewerage monitoring application demonstrate the effectiveness of the proposed framework. Beyond accurate anomaly detection, the standardized edge scores consistently identify the variable relationships responsible for anomalous behavior, providing interpretable insights into the underlying causes of detected deviations. Future work will explore alternative vine structures, such as C-vines, which may provide additional insight into hierarchical and potentially causal relationships among variables by explicitly modeling the influence of central variables on remaining system components. We also plan to extend the framework to temporal and streaming settings, where dependence patterns evolve over time.
\vspace{-.25cm}
\begin{credits}
\subsubsection{\discintname}
The authors have no competing interests to declare that are
relevant to the content of this article. 
\end{credits}
\vspace{-.25cm}
%
%
%
%
\bibliographystyle{splncs04} 
\bibliography{bibliography}

\newpage
\appendix
\renewcommand{\thefigure}{A\arabic{figure}}
\renewcommand{\thetable}{A\arabic{table}}
\setcounter{figure}{0}
\setcounter{table}{0}

\section{Pair-Copula Families}\label{app:families}
This Section of the Appendix contains the closed form copula densities and h-functions of the bivariate families used in our implementation. Each family is parametrized by a parameter vector $\theta$ whose admissible value ranges are reported in Table~\ref{tab:copula-families}. For a bivariate copula $C(u_1,u_2;\theta)$ the copula density is defined as: $$c(u_1,u_2; \theta)=\cfrac{\partial^2C(u_1,u_2;\theta)}{\partial u_1\partial u_2},$$
while the h-functions correspond to the conditional distributions:
$$h_1(u_1,u_2;\theta) = \cfrac{\partial C(u_1,u_2;\theta)}{\partial u_2}, \qquad h_2(u_1,u_2;\theta) = \cfrac{\partial C(u_1,u_2;\theta)}{\partial u_1}.$$
Since $h_2(u_1,u_2;\theta) = h_1(u_2,u_1;\theta)$ for all families considered, we report only the first h-function $h_1$.

\begin{table}[h]
\centering
\begin{tabular}{lcccc}
\toprule
\textbf{Family} & \textbf{1st Parameter} & \textbf{2nd Parameter} & \textbf{Lower TD} & \textbf{Upper TD} \\
\midrule
Gaussian   & $\rho \in [-1, 1]$  & --                  & 0 & 0 \\
Student-t  & $\rho \in [-1, 1]$  & $\nu \in[2,30]$           & + & + \\
Clayton    & $\delta \in  [0, 7.5]$        & --                  & + & 0 \\
Frank      & $\theta \in [-10, 10] \backslash 0$     & --                  & 0 & 0 \\
Gumbel     & $\delta \in [1, 17]$     & --                  & 0 & + \\
Joe        & $\delta \in [1, 10] $     & --                  & 0 & + \\
\bottomrule
\end{tabular}
\caption{Implemented bivariate copula families with their parameter ranges and tail dependence behavior. $+$ indicates non-zero tail dependence; $0$ indicates tail independence.}
\label{tab:copula-families}
\end{table}
\subsection{Gaussian Copula}
\begin{figure}[!t]
    \centering
    \includegraphics[width=\linewidth]{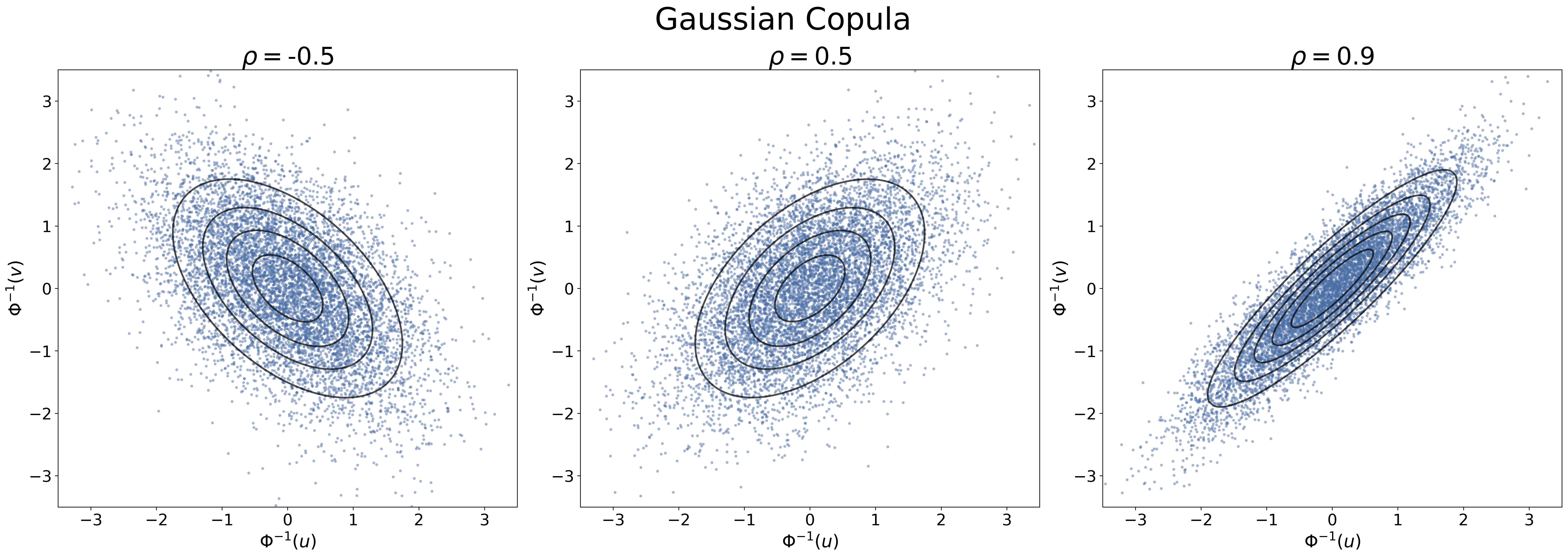}
    \caption{Joint densities induced by the Gaussian copula for different values of $\rho$. All panels use the same scale to ensure comparability.}
    \label{fig:bicopGAUS}
\end{figure}

The Gaussian copula is parametrized by a single parameter, the correlation coefficient $\rho$. It admits values in $\rho\in(-1,1)$, with $\rho = 0$ denoting independence between the two variables. Letting $z_1 = \Phi^{-1}(u_1)$ and $z_2 = \Phi^{-1}(u_2)$, where $\Phi$ denotes the standard normal CDF, the copula density of the Gaussian copula is:
$$
    c(u_1, u_2;\rho) = \frac{1}{\sqrt{1-\rho^2}}\exp{\left(-\frac{\rho^2\left( z_1^2+z_2^2\right)-2\rho z_1z_2}{2\left(1-\rho^2\right)}\right)}
$$
and the h-function is:
$$h_1(u_1,u_2;\rho) = \Phi\!\left( \frac{z_1 - \rho\, z_2}{\sqrt{1-\rho^2}} \right).$$
Fig.~\ref{fig:bicopGAUS} illustrates the dependence structures induced by the Gaussian copula for different values of the correlation coefficient $\rho$. The displayed joint densities are obtained by combining the copula density for the specific parameter configuration with standard Gaussian marginals and are overlaid on samples simulated from the corresponding copula model.

\subsection{Student-t Copula}
\begin{figure}[t!]
    \centering
    \includegraphics[width=\linewidth]{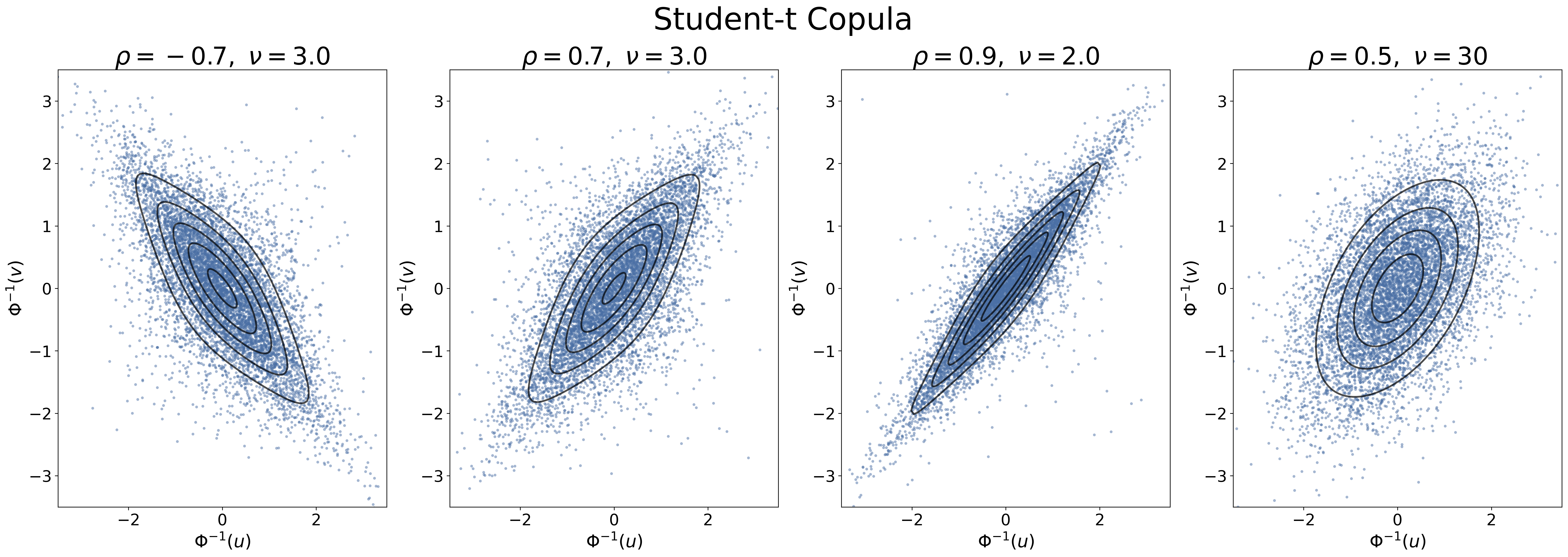}
    \caption{Joint densities induced by the Student-t copula for different values of $\rho$ and $\nu$. All panels use the same scale to ensure comparability.}
    \label{fig:bicopSTUD}
\end{figure}

The Student-t copula is parametrized by two parameters, the correlation coefficient $\rho$ and the degrees of freedom $\nu$. They admit values in $\rho \in(-1,1)$ and $\nu>0$, with smaller values of $\nu$ corresponding to heavier tails and stronger tail dependence, while larger values induce a behavior closer to a Gaussian copula. Letting $z_1 = t_\nu^{-1}(u_1)$ and $z_2 = t_\nu^{-1}(u_2)$, where $t_\nu$ denotes the standard Student-t CDF with $\nu$ degrees of freedom, the copula density is:
$$
c(u_1, u_2; \rho, \nu) = \frac{\Gamma\!\left(\frac{\nu+2}{2}\right)\Gamma\!\left(\frac{\nu}{2}\right)}{\sqrt{1-\rho^2}\;\Gamma\!\left(\frac{\nu+1}{2}\right)^2} \cdot \frac{\left(1 + \dfrac{z_1^2 + z_2^2 - 2\rho\, z_1 z_2}{\nu(1-\rho^2)}\right)^{-\frac{\nu+2}{2}}}{\left(1 + \dfrac{z_1^2}{\nu}\right)^{-\frac{\nu+1}{2}}\left(1 + \dfrac{z_2^2}{\nu}\right)^{-\frac{\nu+1}{2}}}
$$
and the h-function is:
$$
h_1(u_1, u_2; \rho, \nu) = t_{\nu+1}\!\left( \frac{z_1 - \rho\, z_2}{\sqrt{\dfrac{(\nu + z_2^2)(1-\rho^2)}{\nu+1}}} \right).
$$
Fig.~\ref{fig:bicopSTUD} illustrates the dependence structures induced by the Student-t copula for different values of the parameters $\rho$ and $\nu$. The displayed joint densities are obtained by combining the copula density for the specific parameter configurations with standard Gaussian marginals and are overlaid on samples simulated from the corresponding copula model.

\subsection{Clayton Copula}
The Clayton copula is parametrized by a single parameter $\delta$ controlling the strength of dependence, which admits values in $\delta > 0$, with larger values corresponding to stronger lower-tail dependence. The copula density is:
$$
    c(u_1, u_2;\delta) = (1+\delta)\,(u_1 u_2)^{-1-\delta} \left( u_1^{-\delta} + u_2^{-\delta} - 1 \right)^{-1/\delta - 2},
$$
and the h-function is:
$$
    h_1(u_1, u_2;\delta) = u_2^{-\delta - 1} \left( u_1^{-\delta} + u_2^{-\delta} - 1 \right)^{-1 - 1/\delta}.
$$
\begin{figure}[b!]
    \centering
    \includegraphics[width=\linewidth]{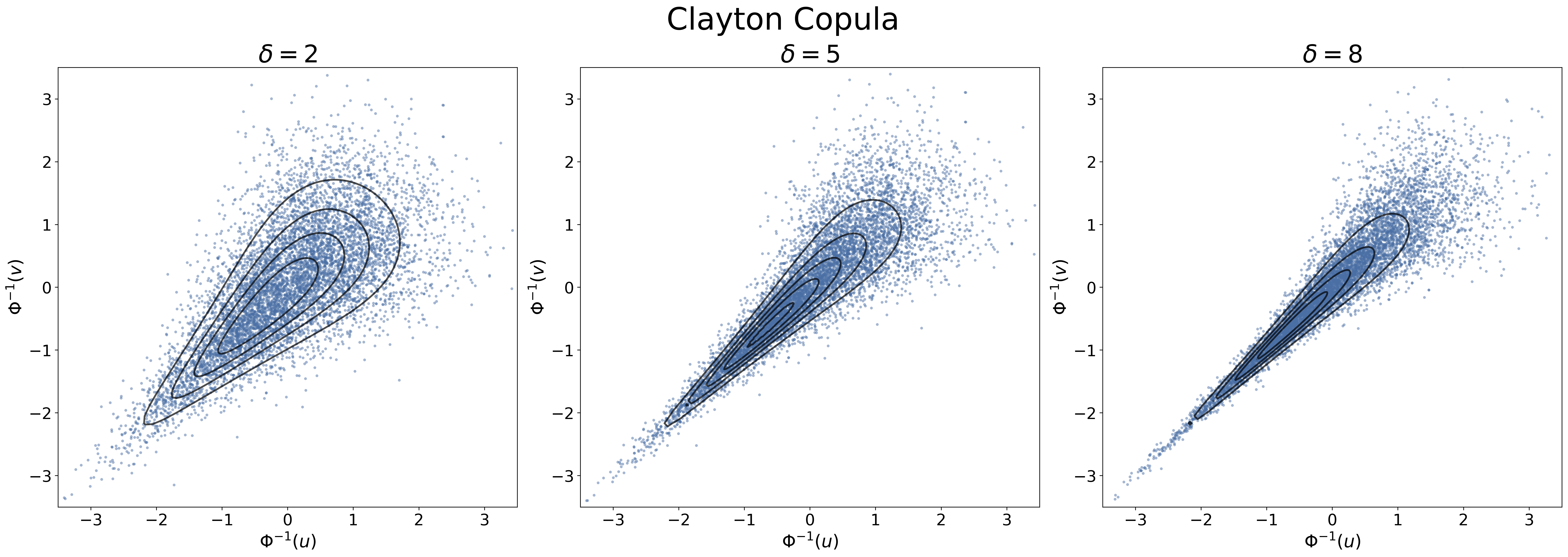}
    \caption{Joint densities induced by the Clayton copula for different values of $\delta$. All panels use the same scale to ensure comparability.}
    \label{fig:bicopCLAY}
\end{figure}
Fig.~\ref{fig:bicopCLAY} illustrates the dependence structures induced by the Clayton copula for different values of the parameter $\delta$. The displayed joint densities are obtained by combining the copula density for the specific parameter configurations with standard Gaussian marginals and are overlaid on samples simulated from the corresponding copula model.

\subsection{Frank Copula}
\begin{figure}[t!]
    \centering
    \includegraphics[width=\linewidth]{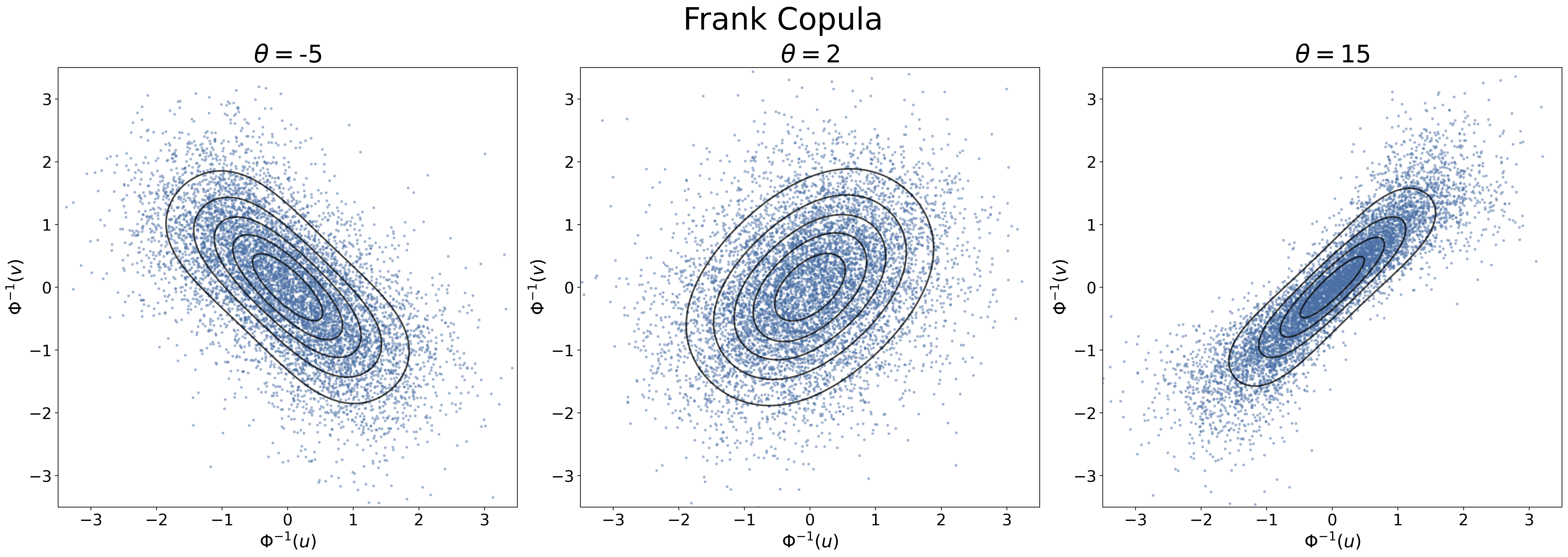}
    \caption{Joint densities induced by the Frank copula for different values of $\theta$. All panels use the same scale to ensure comparability.}
    \label{fig:bicopFRANK}
\end{figure}

The Frank copula is parametrized by a single parameter $\theta$, controlling the strength and direction of dependence, which admits values in $\theta \neq 0$; positive values correspond to positive dependence and negative values to negative dependence, with the independence copula recovered as $\theta \to 0$. The copula density is:
$$
    c(u_1, u_2; \theta) = \cfrac{\theta\left(1-e^{-\theta}\right) e^{-\theta(u_1 + u_2)}}{\left[\left(1-e^{-\theta}\right) - \left(1-e^{-\theta u_1}\right)\left(1-e^{-\theta u_2}\right)\right]^2},
$$
and the h-function is:
$$
    h_1(u_1, u_2; \theta) = \cfrac{e^{-\theta u_2}\left(1 - e^{-\theta u_1}\right)}{e^{-\theta u_1} + e^{-\theta u_2} - e^{-\theta(u_1 + u_2)} - e^{-\theta}}.
$$
Fig.~\ref{fig:bicopFRANK} illustrates the dependence structures induced by the Frank copula for different values of the parameter $\theta$. The displayed joint densities are obtained by combining the copula density for the specific parameter configurations with standard Gaussian marginals and are overlaid on samples simulated from the corresponding copula model.

\subsection{Gumbel Copula}
\begin{figure}[t!]
    \centering
    \includegraphics[width=\linewidth]{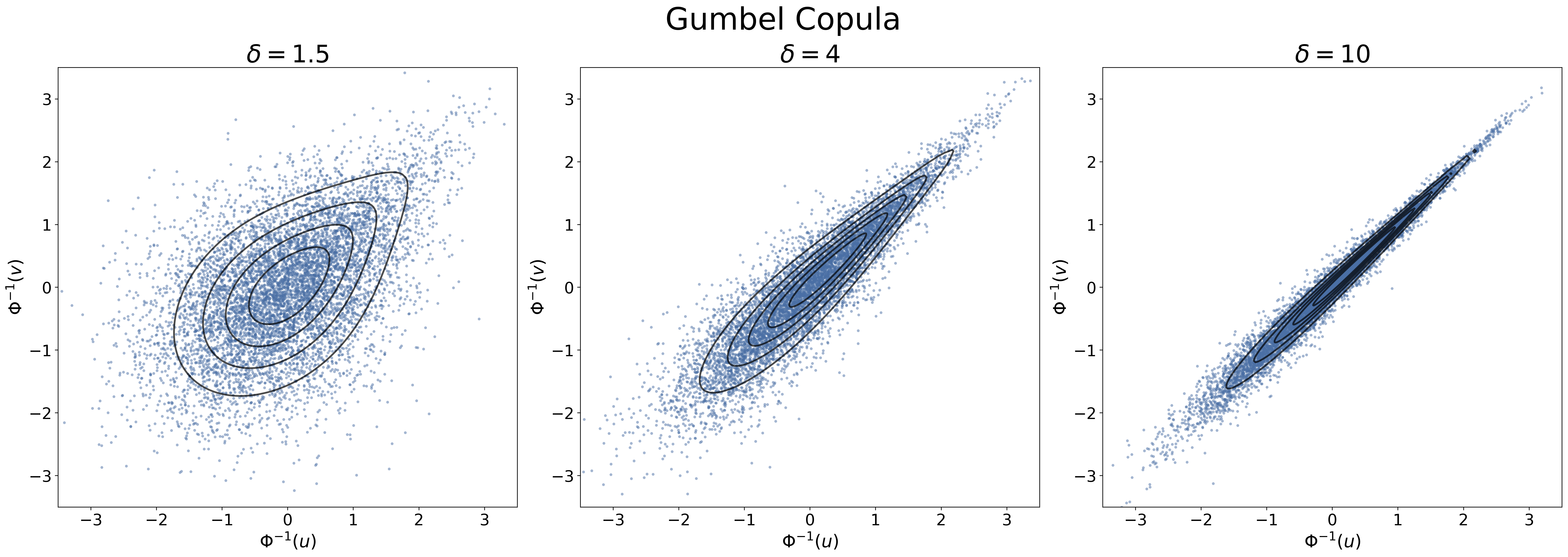}
    \caption{Joint densities induced by the Gumbel copula for different values of $\delta$. All panels use the same scale to ensure comparability.}
    \label{fig:bicopGUMB}
\end{figure}

The Gumbel copula is parametrized by a single parameter $\delta$, controlling the strength of dependence, which admits values in $\delta \geq 1$, with larger values corresponding to stronger upper-tail dependence and $\delta = 1$ recovering independence. With $C(u_1, u_2; \delta) = \exp\!\left[-\left\{(-\log u_1)^{\delta} + (-\log u_2)^{\delta}\right\}^{1/\delta}\right]$ the density is:
\begin{align}
    c(u_1,u_2;\delta) &= C(u_1, u_2; \delta)\cdot(u_1 u_2)^{-1} \nonumber\\
    &\quad \times \left\{(-\log u_1)^{\delta} + (-\log u_2)^{\delta}\right\}^{1/\delta - 2} \nonumber\\
    &\quad \times \left[(-\log u_1)(-\log u_2)\right]^{\delta-1} \nonumber\\
    &\quad \times \left[\left\{(-\log u_1)^{\delta} + (-\log u_2)^{\delta}\right\}^{1/\delta} + \delta - 1\right],
    \nonumber
\end{align}
and the h-function is:
$$
    h_1(u_1, u_2; \delta) = C(u_1, u_2; \delta)\cdot\frac{1}{u_2}\cdot(-\log u_2)^{\delta-1}\cdot\left\{(-\log u_1)^{\delta} + (-\log u_2)^{\delta}\right\}^{1/\delta - 1}.
$$
Fig.~\ref{fig:bicopGUMB} illustrates the dependence structures induced by the Gumbel copula for different values of the parameter $\delta$. The displayed joint densities are obtained by combining the copula density for the specific parameter configurations with standard Gaussian marginals and are overlaid on samples simulated from the corresponding copula model.

\subsection{Joe Copula}
\begin{figure}[t!]
    \centering
    \includegraphics[width=\linewidth]{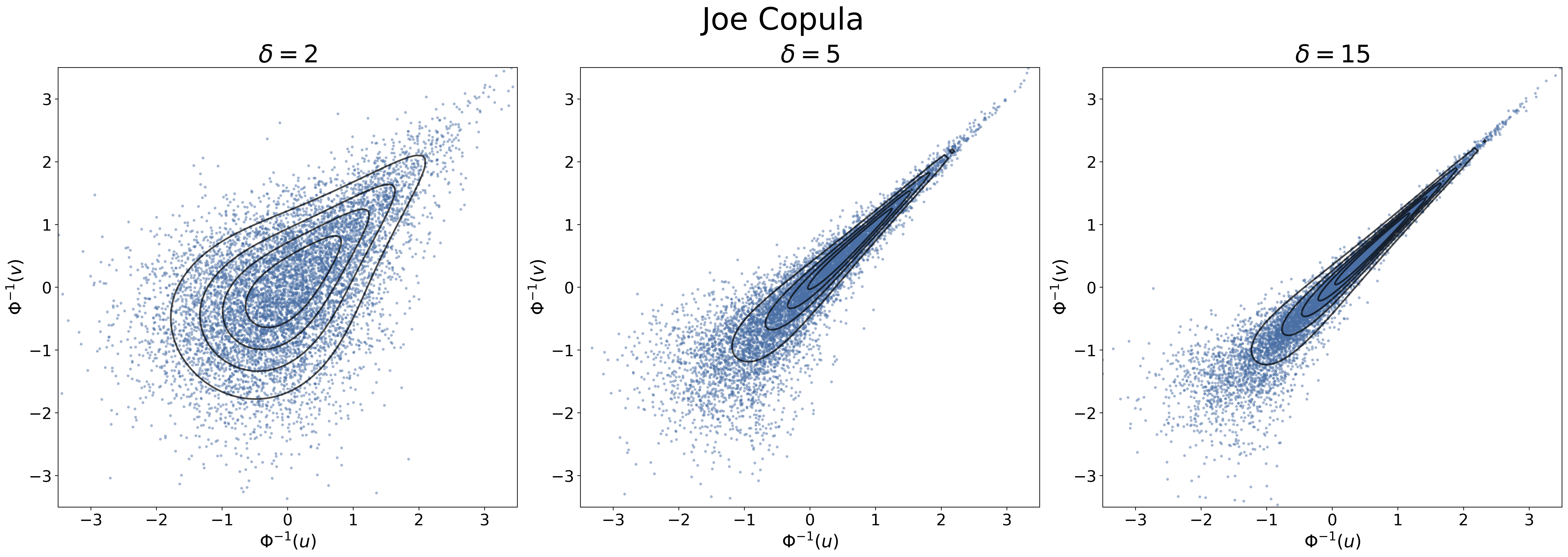}
    \caption{Joint densities induced by the Joe copula for different values of $\delta$. All panels use the same scale to ensure comparability.}
    \label{fig:bicopJOE}
\end{figure}

The Joe copula is parametrized by a single parameter $\delta$, controlling the strength of the dependence, which admits values in $\delta \geq 1$. Larger values correspond to stronger upper-tail dependence. With $t_1 = (1-u_1)^\delta$ and $t_2 = (1-u_2)^\delta$ the copula density can be expressed as:
\begin{align}
    c(u_1,u_2;\delta) &= \left(t_1 + t_2 - t_1 t_2\right)^{1/\delta - 2} \nonumber\\
    &\quad \times \left(\delta - 1 + t_1 + t_2 - t_1 t_2\right) \nonumber\\
    &\quad \times \left[(1-u_1)(1-u_2)\right]^{\delta-1}, \nonumber
\end{align}
and the h-function is:
$$h_1(u_1,u_2;\delta) = \left(1 - t_1\right)\cdot(1-u_2)^{\delta-1}\cdot\left(t_1 + t_2 - t_1 t_2\right)^{1/\delta - 1}.$$
Fig.~\ref{fig:bicopJOE} illustrates the dependence structures induced by the Joe copula for different values of the parameter $\delta$. The displayed joint densities are obtained by combining the copula density for the specific parameter configurations with standard Gaussian marginals and are overlaid on samples simulated from the corresponding copula model.

\subsection{Tail Dependence Coefficients}
\label{subsec:tailDependece}
The dependence structure captured by a bivariate copula can be described through its tail dependence coefficients $\lambda_L$ and $\lambda_U$, which quantify the probability of joint extreme co-movements in the lower and upper tails respectively.
For $U_1, U_2\sim C$, the lower and upper tail dependence coefficients are defined as
\begin{equation}
\label{eq:tailDependence}
    \lambda_L =  \lim_{u\to 0^+}\frac{C(u,u)}{u}, \quad \quad \lambda_U = \lim_{u\to 1^-}\frac{1-2u+C(u,u)}{1-u}\nonumber
\end{equation}
when the limit exists. Both of these coefficients admit values in $[0,1]$, with values close to $0$ indicating tail independence and values close to $1$ indicate strong joint tail behavior. Table~\ref{tab:copula-families} shows the tail dependence behavior for the bivariate families described above.

\newpage
\section{Datasets}\label{app:data}
\renewcommand{\thefigure}{B\arabic{figure}}
\renewcommand{\thetable}{B\arabic{table}}
\setcounter{figure}{0}
\setcounter{table}{0}
This Section provides supplementary graphical representations of the datasets analyzed in Section~\ref{sec:experiments}, which were omitted from the main body due to space constraints.
Fig.~\ref{fig:WiltOV} shows the pairwise scatterplots for every pair of features included in the Wilt dataset. The ordinary and anomalous observations overlap almost perfectly across all variable pairs, suggesting that the anomalies do not arise from extreme marginal behaviors, but from different dependence patterns that exist for ordinary and anomalous observations, as is clearly visible in Fig.~\ref{fig:WiltPSO}. \begin{figure}[H]
    \centering
    \includegraphics[width=1\linewidth]{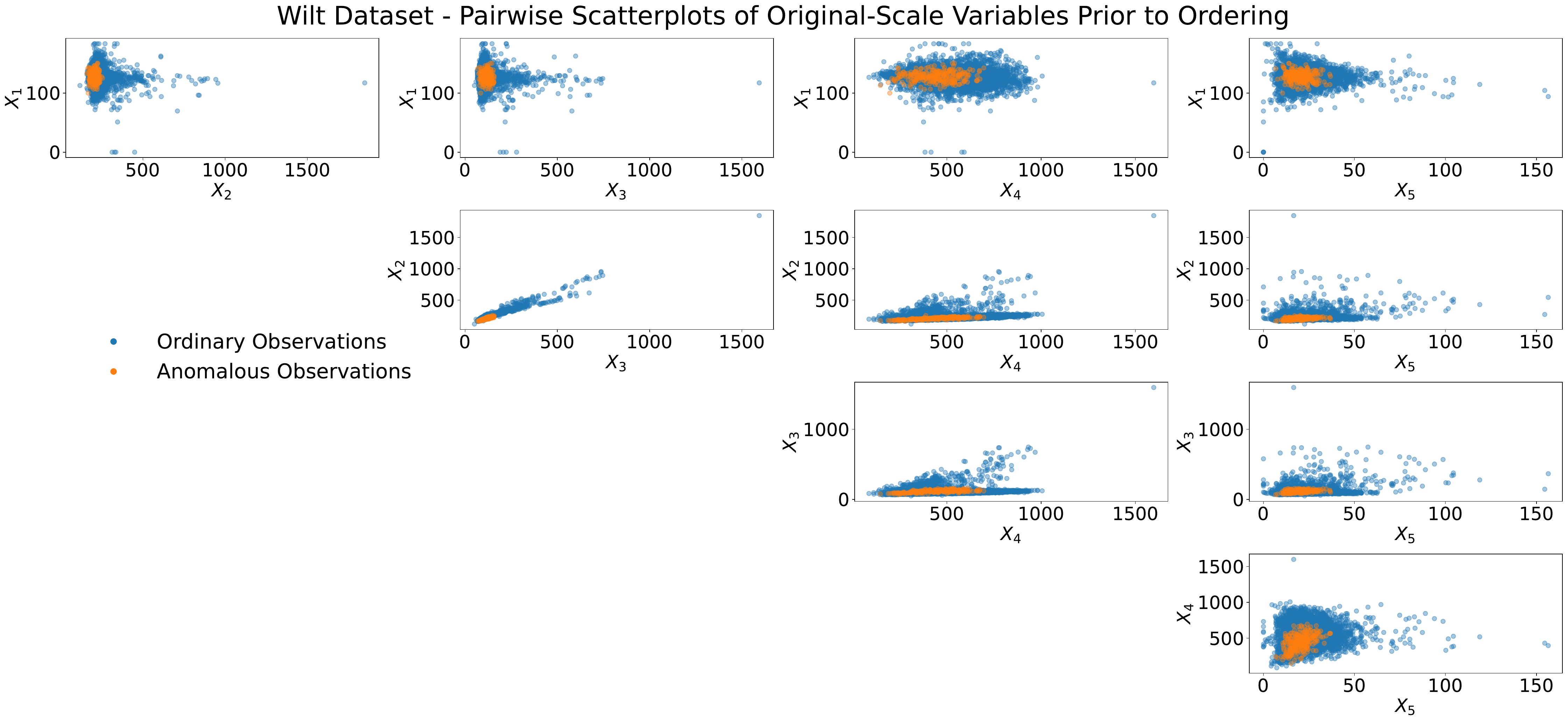}
    \caption{\textbf{Wilt:} Pairwise scatterplots of the variables in their original scale. Ordinary and anomalous observations are included and coloured based on their label.}
    \label{fig:WiltOV}
\end{figure}
\begin{figure}[!b]
\centering
\includegraphics[width=1\linewidth]{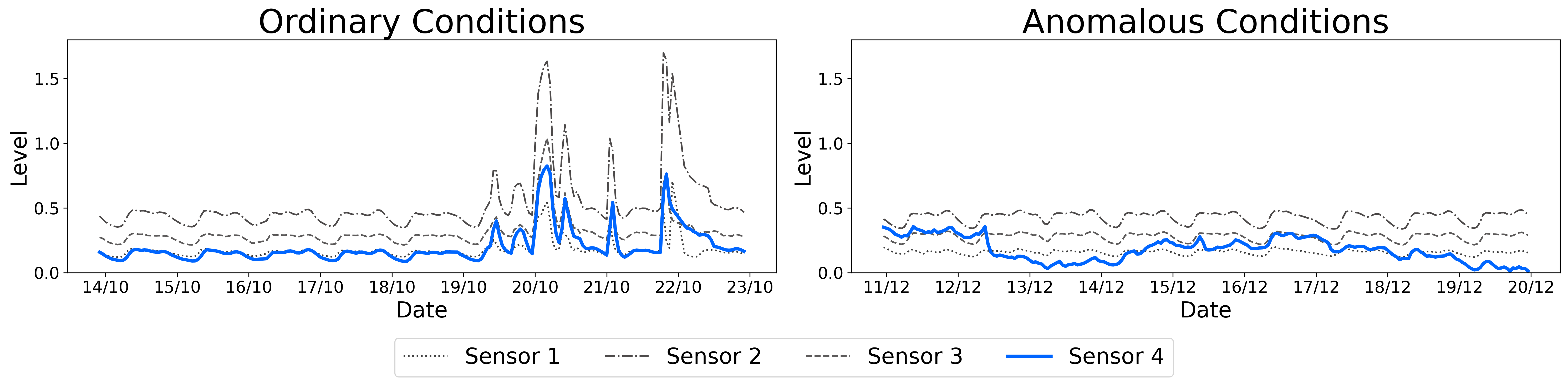}
\caption{\textbf{Sewerage:} Hourly level recordings for the four sensors. Left: ordinary behavior, with coherent daily patterns in dry conditions and simultaneous rainfall spikes. Right: the anomalous state of Sensor 4, whose readings no longer follow the shared network behavior.}
\label{fig:BrianzaTS}
\end{figure}

Fig.~\ref{fig:BrianzaTS} illustrates the multivariate time series from the Sewerage dataset, showing hourly readings for Sensors 1–4. The left panel depicts ordinary conditions, characterized by regular daily patterns and spikes during rainfall events. The right panel shows anomalous behavior in Sensor 4 (highlighted in blue), which breaks from the shared dynamics of the network while staying within its operational range.

\begin{figure}[t!]
    \centering
    \includegraphics[width=1\linewidth]{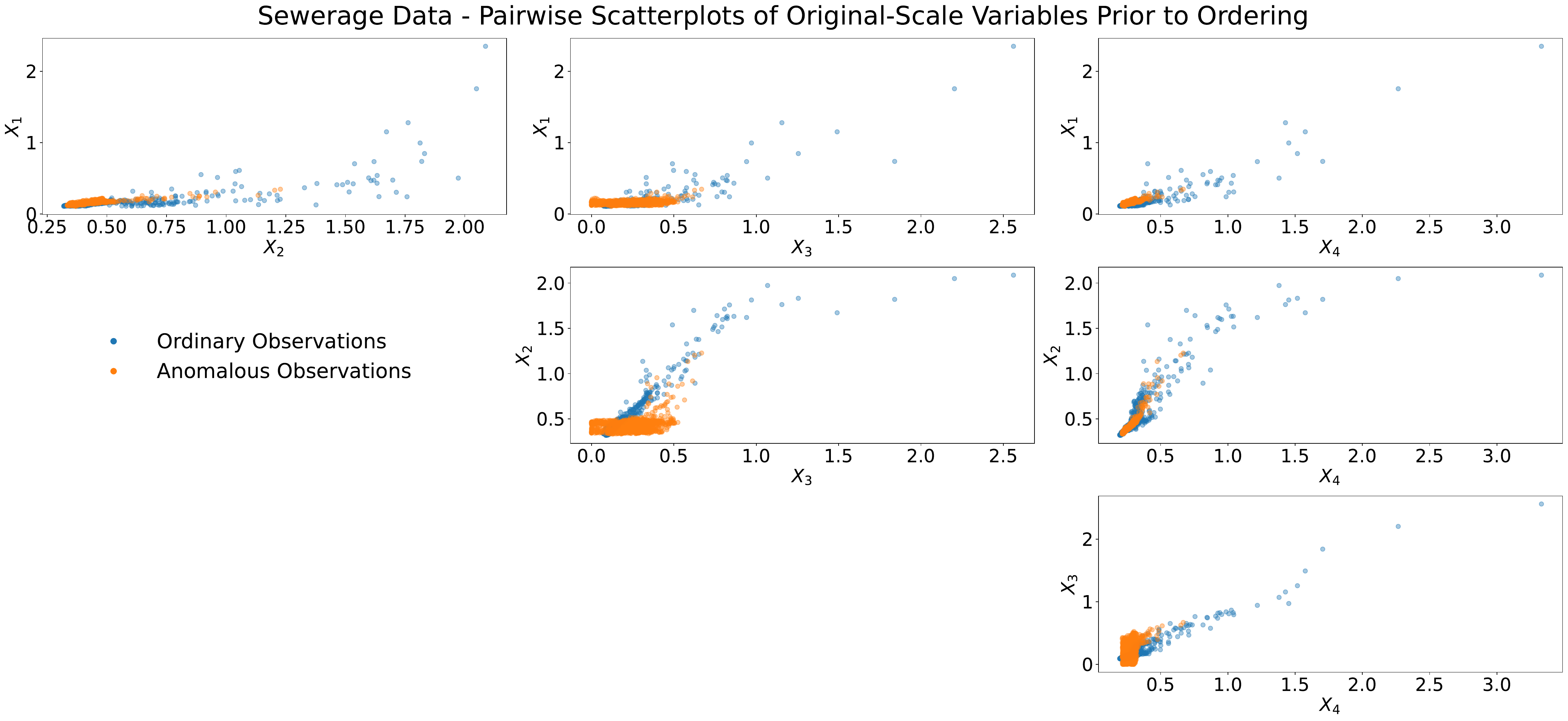}
    \caption{\textbf{Sewerage:} Pairwise scatterplots of the variables in their original scale. Ordinary and anomalous observations are included and coloured based on their label.}
    \label{fig:brianzaOV}
\end{figure}

\begin{figure}[!b]
    \centering
    \includegraphics[width=1\linewidth]{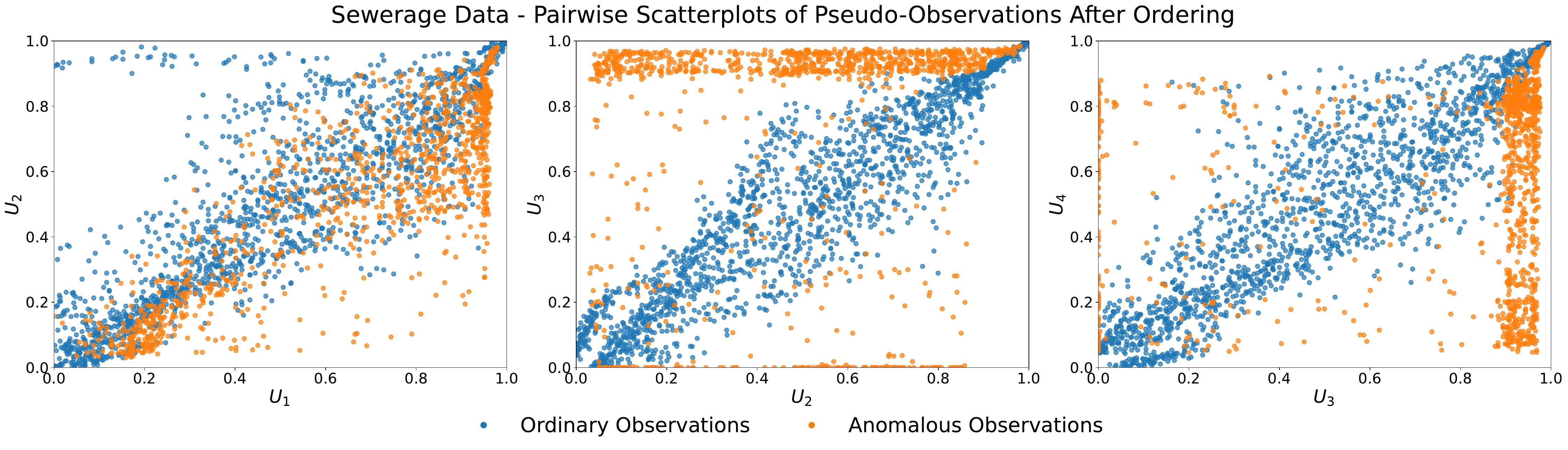}
    \caption{\textbf{Sewerage:} Pairwise scatterplots of the ordered pseudo-observations. Variables pairs plotted together form the inputs to the pair-copulas at the first tree $T_1$.}
    \label{fig:brianzaPSO}
\end{figure}

Fig.~\ref{fig:brianzaOV} shows the pairwise scatterplots for each pair of features included in the real-world Sewerage Dataset. In this scenario the ordinary and anomalous observations overlap significantly for most variable pairs with the exception of those involving feature $X_3$, the one containing the reading from the anomalous Sensor 4. This pattern is further confirmed by Fig.~\ref{fig:brianzaPSO}, where the clearest separations between ordinary and anomalous behavior can be attributed to the same feature. Notably, the second and third panels represent the pseudo-observations which form the inputs of edges $(1,2)$ and $(1,3)$ of the D-vine, whose relevance in identifying anomalies was described in Section~\ref{sec:experiments}.

\newpage
\section{Additional Results}\label{app:results}
\renewcommand{\thefigure}{C\arabic{figure}}
\renewcommand{\thetable}{C\arabic{table}}
\setcounter{figure}{0}
\setcounter{table}{0}
This section presents additional experimental results for both datasets analyzed in Section~\ref{sec:experiments}, supplementing those reported in the main body.
\subsection{Wilt Dataset}
\label{app:wilt}
\begin{figure}[!b]
    \centering
    \includegraphics[width=1\linewidth]{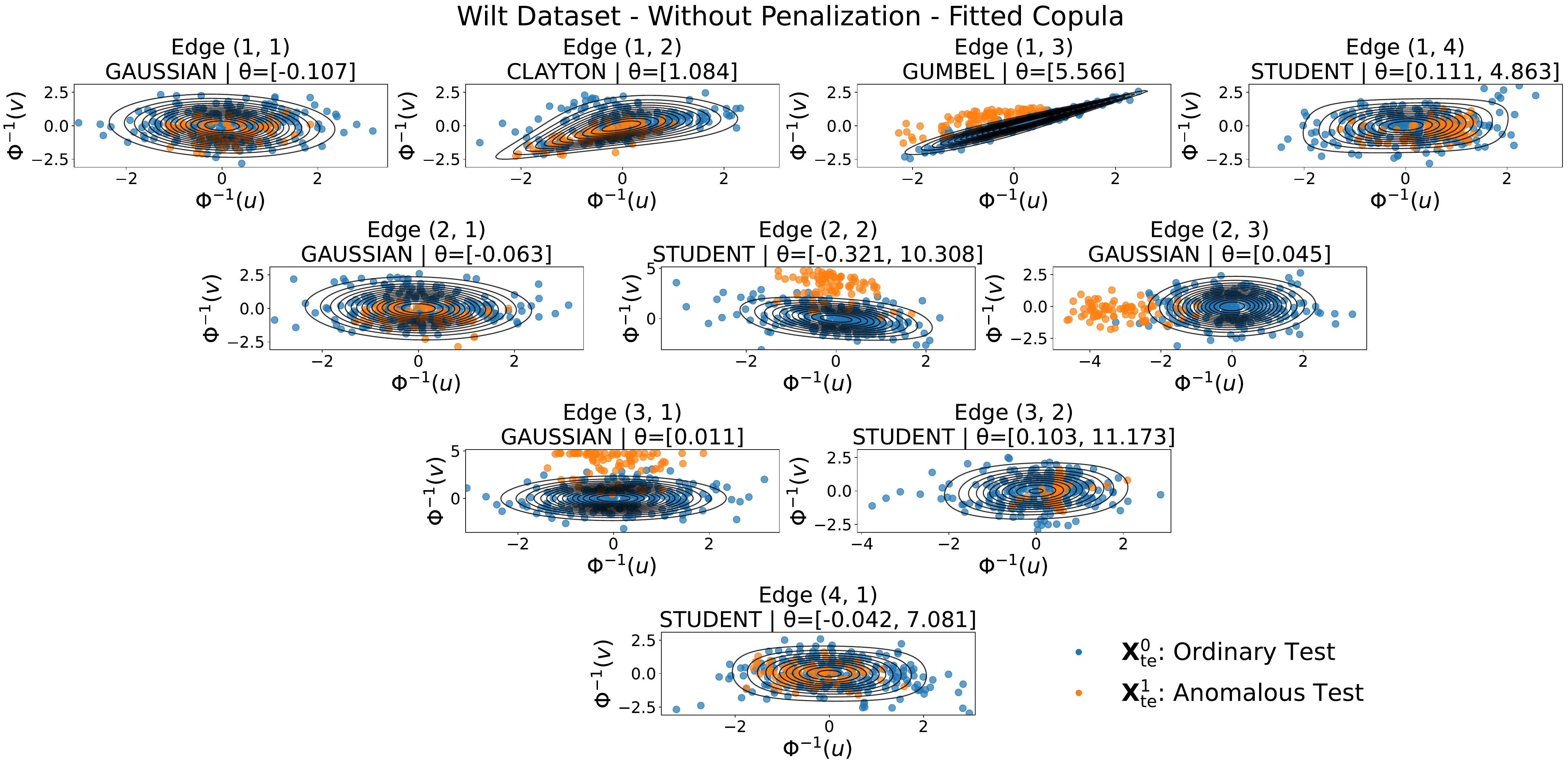}
    \caption{\textbf{Wilt:} Fitted pair-copulas of the D-vine without penalization, shown as joint densities. Each panel can be read in isolation, as the dependence accumulated in earlier trees is embedded in the inputs of each panel through the h-function cascade.}
    \label{fig:wiltFitNoPen}
\end{figure}

Fig.~\ref{fig:wiltFitNoPen} shows the D-vine configuration estimated by the fitting procedure without penalization. Each panel visualizes the joint density induced by the fitted pair-copula combined with standard Gaussian marginals. When compared to Fig.~\ref{fig:WiltedFittedYesPen}, most edges maintain similar family-parameter configurations when fitting with and without penalization, with the exceptions being edges $(1,3)$ and $(2,2)$. Given that these edges are the ones associated with pairs of variables which present a clearer class separation, the penalized version of the fitting procedure is better able to discriminate between the two classes, forcing a configuration which assigns lower density to anomalous observations.

\begin{figure}[!t]
    \centering
    \includegraphics[width=1\linewidth]{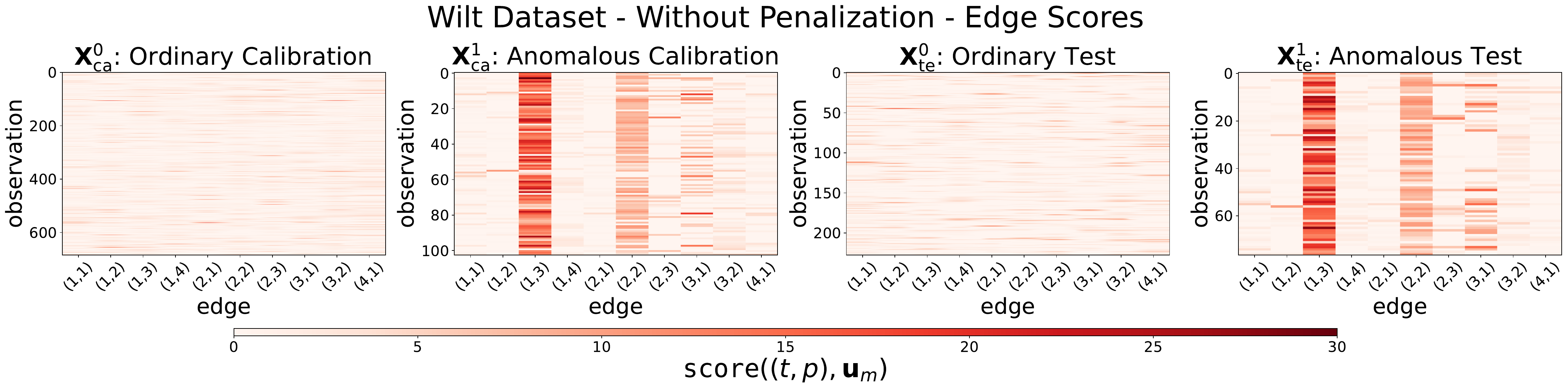}
 \caption{\textbf{Wilt:} Standardized edge scores $\mathtt{score}\big((t,p), \mathbf{u}_m\big)$ for the Wilt dataset across the calibration and test sets, for fitting without penalization.}
    \label{fig:wiltAttributionNoPen}
\end{figure}

This factor is also visible from Fig.~\ref{fig:wiltAttributionNoPen}, which highlights the standardized edge scores across the calibration and test sets, for both ordinary and anomalous observations. Similarly to Fig.~\ref{fig:wiltedAttribution}, edges $(1,3)$ and $(2,2)$ present higher scores for anomalous observations compared to other edges. However, when comparing with the penalized version, the scores associated to these edges have values lower in magnitude, especially for edge $(2,2)$.

\begin{figure}[!t]
    \centering
    \includegraphics[width=1\linewidth]{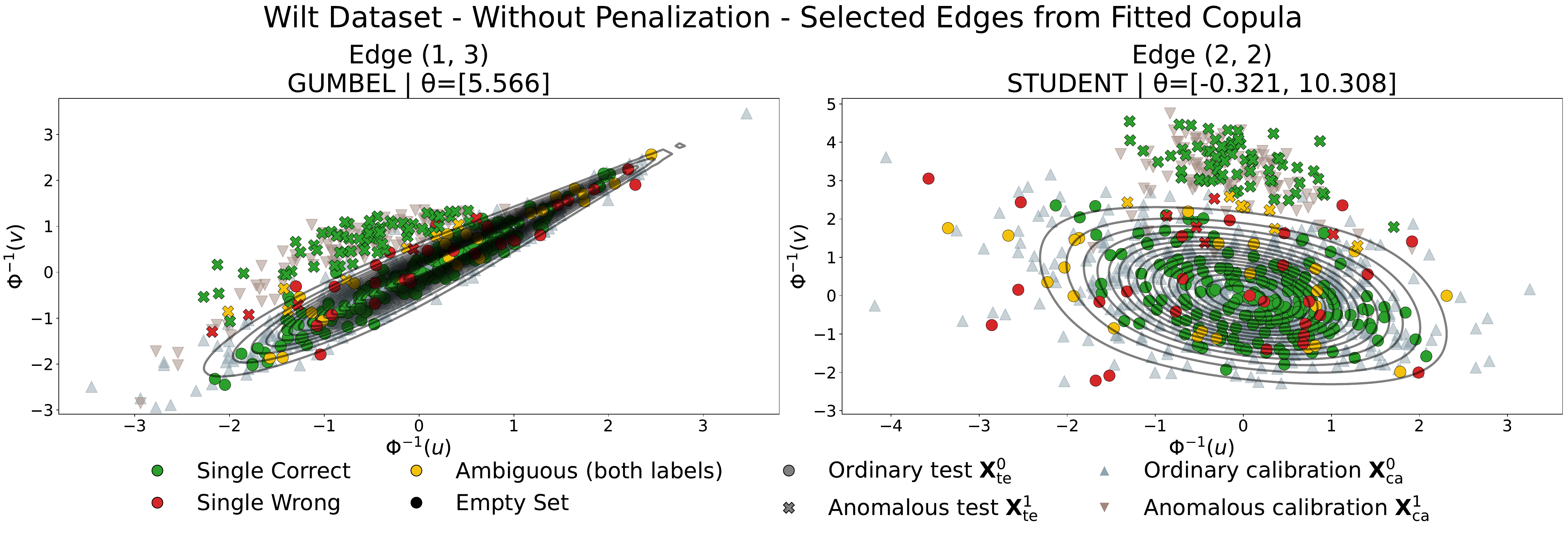}
 \caption{\textbf{Wilt:} Anomaly detection results under non-penalized fitting at edges $(1,3)$ (left) and $(2,2)$ (right), the two most relevant for anomaly detection. Marker color indicates prediction correctness; shape distinguishes ordinary from anomalous observations.}
    \label{fig:wiltedAD_nopen}
\end{figure}

Lastly, Fig.~\ref{fig:wiltedAD_nopen} provides a graphical representation showing the classification obtained at these edges.

The results are also presented in numerical form in~\ref{tab:wilt-nopencp}.

\newcolumntype{Y}{>{\centering\arraybackslash}X}
\begin{table}[H]
\centering
\renewcommand{\arraystretch}{1.2}
\begin{tabularx}{\textwidth}{c|YYYYYYY}
\textbf{Dataset} &
\textbf{True Class} &
$\mathbf{n}$ &
\textbf{Coverage} &
\textbf{Single Correct} &
\textbf{Single Wrong} &
\textbf{Both Labels} &
\textbf{Empty Set} \\ \hline
\multirow{2}{*}{\begin{tabular}[c]{@{}c@{}}Wilt\\ Dataset\end{tabular}}
& Ordinary  & 228  & 0.886 & 179 & 26 & 23 & 0 \\ \cline{2-8}
& Anomalous &  77 &  0.935 & 65 & 5 & 7 & 0\\ \hline
\multirow{2}{*}{\begin{tabular}[c]{@{}c@{}}Sewerage\\ Data\end{tabular}}
& Ordinary  & 256 & 0.887 & 116 & 29 & 111 & 0 \\ \cline{2-8}
& Anomalous & 335 & 0.925 & 245 & 23 & 67 & 0 \\ \hline
\end{tabularx}
\caption{Class-conditional conformal prediction results on the Wilt and Sewerage data test sets at $\alpha = 0.1$, D-vine fitted without penalization. Coverage is the fraction of observations whose prediction region contains the true label (target $\geq 0.9$). The remaining columns report the composition of the prediction regions.}

\label{tab:wilt-nopencp}
\end{table}

\subsection{Sewerage Data}
\label{app:sewerage}

\begin{figure}[H]
    \centering
    \includegraphics[width=1\linewidth]{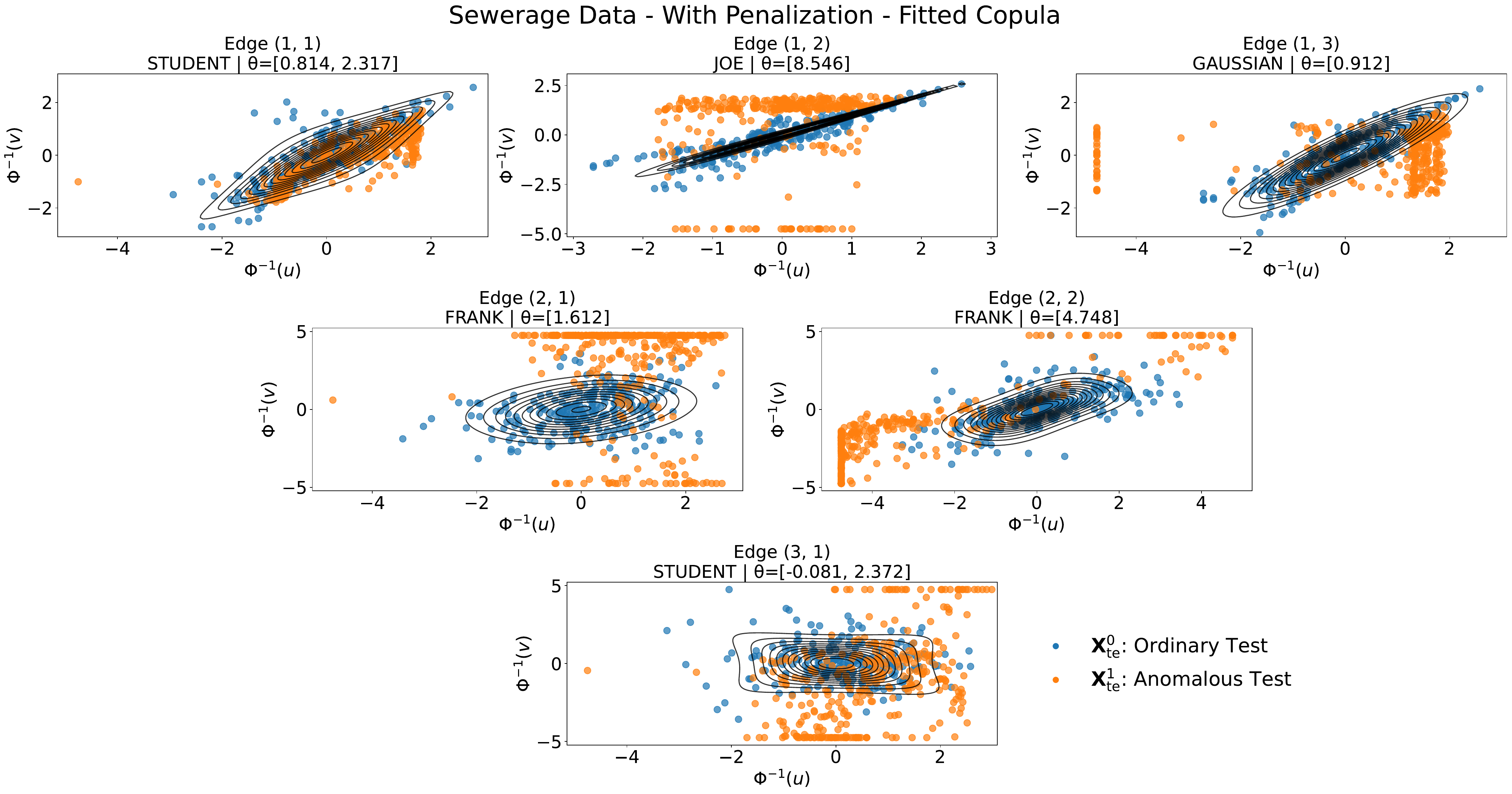}
    \caption{\textbf{Sewerage:} Fitted pair-copulas of the D-vine with penalization, shown as joint densities. Each panel can be read in isolation, as the dependence accumulated in earlier trees is embedded in the inputs of each panel through the h-function cascade.}
    \label{fig:brianzaFitPen}
\end{figure}

\begin{figure}[!b]
    \centering
    \includegraphics[width=1\linewidth]{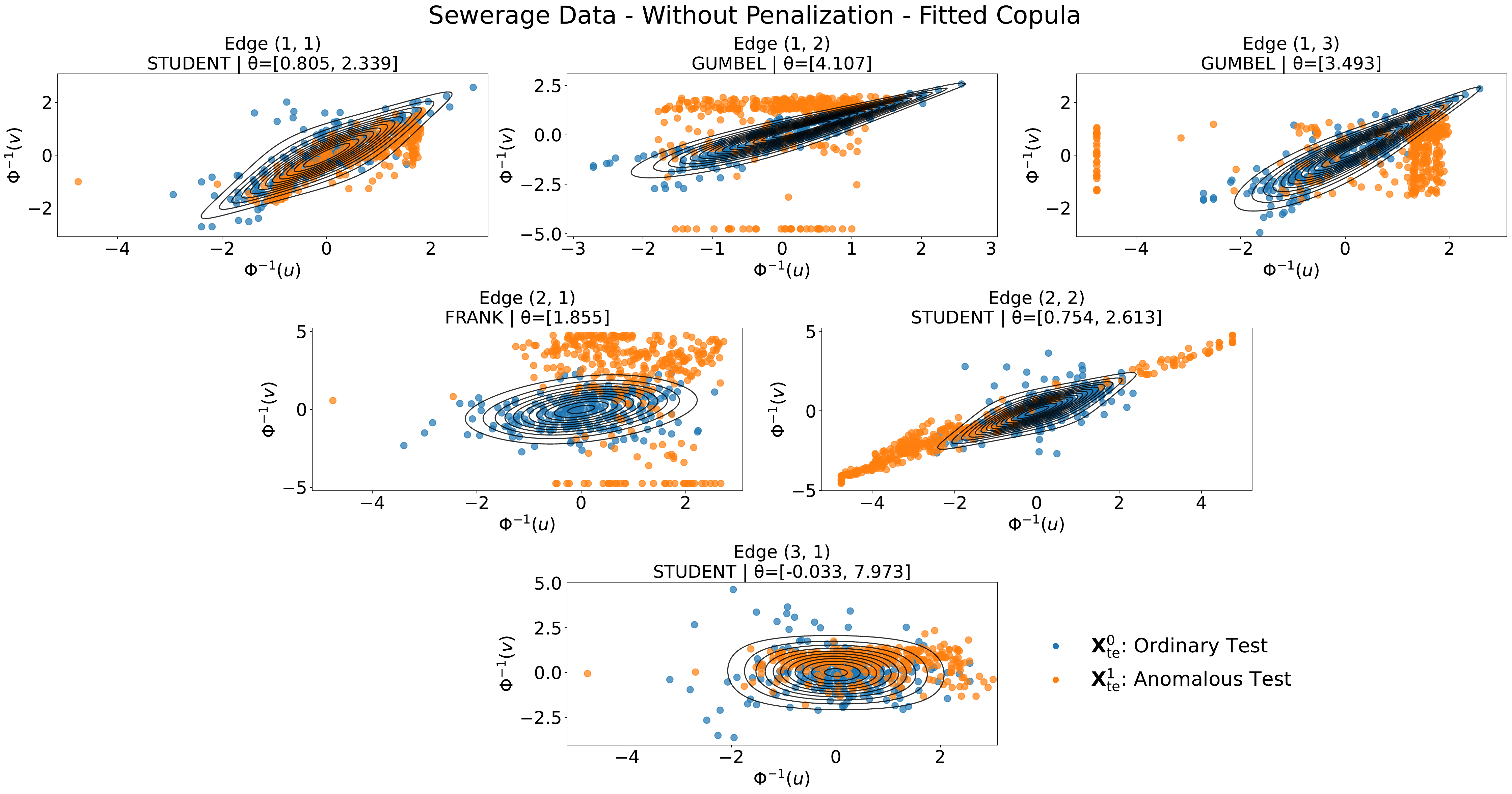}
    \caption{\textbf{Sewerage:} Fitted pair-copulas of the D-vine without penalization, shown as joint densities. Each panel can be read in isolation, as the dependence accumulated in earlier trees is embedded in the inputs of each panel through the h-function cascade.}
    \label{fig:brianzaFitNoPen}
\end{figure}

Similarly to the Wilt dataset, the most notable differences in configurations between the penalized and non penalized versions can be observed for the edges which present a clearer separation between ordinary and anomalous observations. When comparing Figures~\ref{fig:brianzaFitPen} and~\ref{fig:brianzaFitNoPen}, the most notable differences can be observed in edges $(1,2)$ and $(1,3)$, which are the edges in $T_1$ responsible for modeling the features associated with the anomalous Sensor 4. 

\begin{figure}[!b]
    \centering
    \includegraphics[width=1\linewidth]{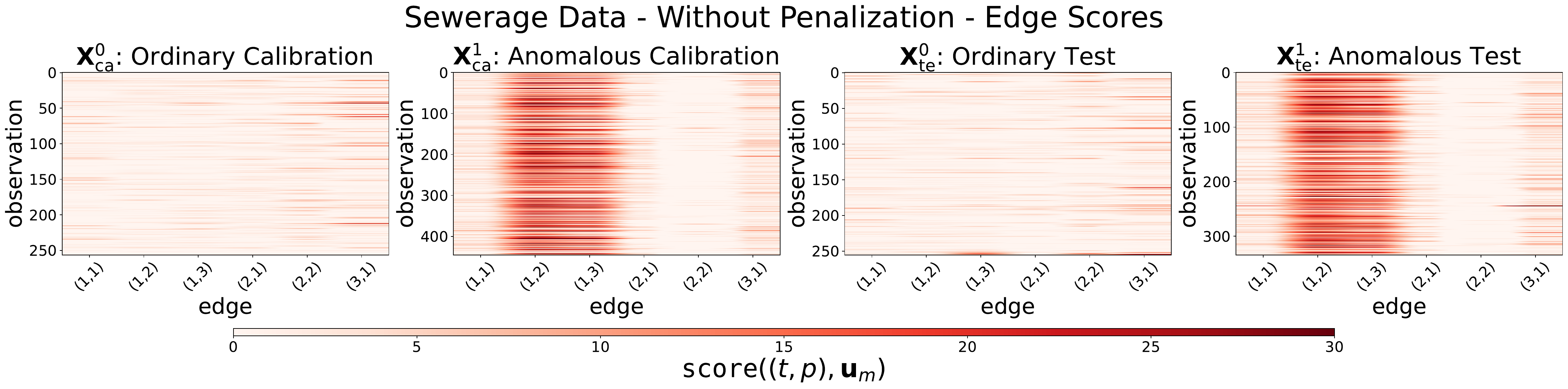}
 \caption{\textbf{Sewerage:} Standardized edge scores $\mathtt{score}\big((t,p), \mathbf{u}_m\big)$ for the Sewerage dataset across the calibration and test sets, for fitting without penalization.}
    \label{fig:brianzaAttributionNoPen}
\end{figure}

Fig.~\ref{fig:brianzaAttributionNoPen} shows the standardized edge scores for the fitting procedure without penalization. When comparing with Fig.~\ref{fig:brianzaAttribution}, edges $(1,2)$, $(1,3)$ and $(3,1)$ present higher than average score values but the magnitude is more muted.

\begin{figure}[!b]
    \centering
    \includegraphics[width=1\linewidth]{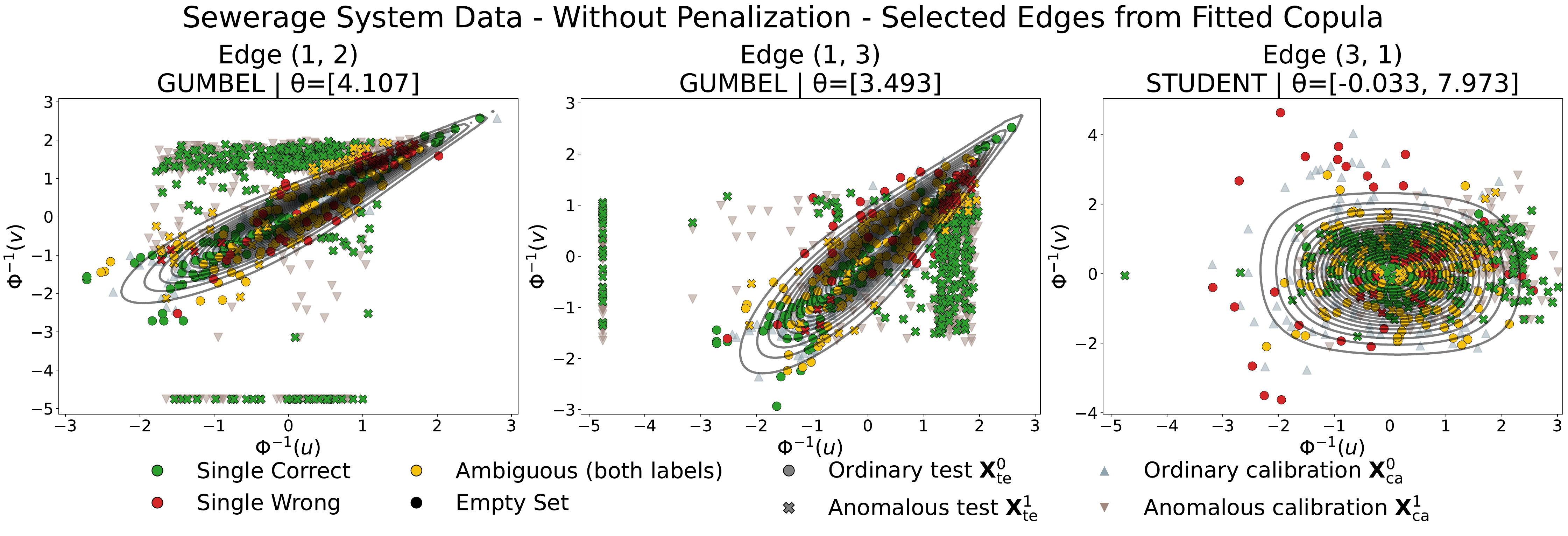}
 \caption{\textbf{Sewerage:} Anomaly detection results under non-penalized fitting at edges $(1,2)$, $(1,3)$ and $(3,1)$, the three most relevant for anomaly detection. Marker color indicates prediction correctness; shape distinguishes ordinary from anomalous observations.}
    \label{fig:brianzaAD_nopen}
\end{figure}

Lastly, Fig.~\ref{fig:brianzaAD_nopen} provides a graphical representation showing the classification obtained at these edges. Notably, when comparing with Fig.~\ref{fig:brianzaSelectedEdges}, a higher number of observations are assigned both labels as can be seen by the prevalence of yellow-colored observations in all represented edges. This result can also be appreciated from a numerical standpoint in Table~\ref{tab:wilt-nopencp}, where the number of observation for the Sewerage datasets which are assigned both labels is significantly higher than its penalized counterpart in Table~\ref{tab:wilt-cp}. For this dataset, the omission of a penalization term manifests with less confident prediction in the overlap regions between the two classes, resulting in more conservative predictions.

\end{document}